\documentclass[10pt,journal,compsoc]{IEEEtran}

\ifCLASSOPTIONcompsoc
  \usepackage[nocompress]{cite}
\else
  \usepackage{cite}
\fi
\usepackage{ragged2e}
\ifCLASSINFOpdf
  \usepackage[pdftex]{graphicx}

\else

\fi

\usepackage{comment}
\usepackage{xcolor}
\usepackage{amsmath}         
\usepackage{amssymb}         
\usepackage{graphicx}        
\usepackage{booktabs}        
\usepackage{threeparttable}  
\usepackage{array}           
\usepackage{multirow}        
\usepackage{mathtools}       

\definecolor{darkblue}{rgb}{0.0,0.0,0.5}
\definecolor{darkred}{rgb}{0.5,0.0,0.0}
\usepackage[
    colorlinks=true,
    linkcolor=darkblue,
    citecolor=darkblue,
    urlcolor=darkblue
]{hyperref}
\usepackage{algorithm}
\usepackage{tikz}
\usepackage{multirow}
\usepackage[edges]{forest}
\usetikzlibrary{arrows.meta,positioning,shapes,trees}
\usepackage{algpseudocode}
\usepackage{booktabs}
\usepackage{amsthm}
\newtheorem{definition}{Definition}

\definecolor{hidden-draw}{RGB}{20,68,106}
\definecolor{hidden-pink}{RGB}{255,245,247}
\definecolor{hidden-red}{RGB}{180,0,0}
\definecolor{hiddendraw}{RGB}{20,68,106}
\definecolor{mygreen}{rgb}{0.2,0.6,0.4}
\definecolor{output-white}{rgb}{1,1,1}
\definecolor{output-black}{rgb}{0,0,0}

\usepackage{url}

\begin{document}
\title{Hyperbolic Graph Neural Networks: A Review of Methods and Applications}

\author{
Menglin Yang, Min Zhou, Tong Zhang, Jiahong Liu, Zhihao Li, Lujia Pan, Hui Xiong, Irwin King
\IEEEcompsocitemizethanks{
\IEEEcompsocthanksitem Menglin Yang, Zhihao Li, and Hui Xiong are with Hong Kong University of Science and Technology (Guangzhou). Email: \{menglinyang@,zli416@connect., xionghui@\}hkust-gz.edu.cn,  
\IEEEcompsocthanksitem Min Zhou and Lujia Pan are with Huawei Technologies Co., Ltd. Email: \{zhoumin27,panlujia\}@huawei.com
\IEEEcompsocthanksitem Jiahong Liu and Irwin King are with The Chinese University of Hong Kong. Email:{jiahong.liu21@gmail.com, king@cse.cuhk.edu.hk}
\IEEEcompsocthanksitem Tong Zhang is with Zhejiang University. Email:hzzhangtong@foxmail.com.
}
\thanks{Corresponding author: Min Zhou}
\thanks{\textcolor{blue}{Note — The latest draft was circulated under the title “Hyperbolic Graph Learning: A Comprehensive Review.” The present arXiv version retains the original title, “Hyperbolic Graph Neural Networks: A Review of Methods and Applications,” for consistency, while incorporating substantial revisions and extensions.}}
}

\markboth{}%
{Shell \MakeLowercase{\textit{et al.}}: Bare Advanced Demo of IEEEtran.cls for IEEE Computer Society Journals}

\IEEEtitleabstractindextext{%
\justify
\begin{abstract}
Graph representation learning in Euclidean space, despite its widespread adoption and proven utility in many domains, often struggles to effectively capture the inherent hierarchical and complex relational structures prevalent in real-world data, particularly for datasets exhibiting a highly non-Euclidean latent anatomy or power-law distributions. Hyperbolic geometry, with its constant negative curvature and exponential growth property, naturally accommodates such structures, offering a promising alternative for learning rich graph representations. This survey paper provides a comprehensive review of the rapidly evolving field of Hyperbolic Graph Learning (HGL). 
We systematically categorize and analyze existing methods broadly dividing them into (1) hyperbolic graph embedding-based techniques, (2) graph neural network-based hyperbolic models, and (3) emerging paradigms.
Beyond methodologies, we extensively discuss diverse applications of HGL across multiple domains, including recommender systems, knowledge graphs, bioinformatics, and other relevant scenarios, demonstrating the broad applicability and effectiveness of hyperbolic geometry in real-world graph learning tasks. 
Most importantly, we identify several key challenges that serve as directions for advancing HGL, including handling complex data structures, developing geometry-aware learning objectives, ensuring trustworthy and scalable implementations, and integrating with foundation models, e.g., large language models. We highlight promising research opportunities in this exciting interdisciplinary area. A comprehensive repository can be found at \url{https://github.com/digailab/awesome-hyperbolic-graph-learning}
\end{abstract}

\begin{IEEEkeywords}
Complex Data, Hyperbolic Geometry, Graph Embedding, Graph Neural Network, Graph Representation Learning.
\end{IEEEkeywords}}

\maketitle
\IEEEdisplaynontitleabstractindextext
\IEEEpeerreviewmaketitle

\ifCLASSOPTIONcompsoc
\IEEEraisesectionheading{\section{Introduction}\label{sec:introduction}}
\else
\section{Introduction}
\label{sec:introduction}
\fi

\IEEEPARstart{G}{raphs} are fundamental data structures that pervade real-world complex systems, including social networks~\cite{grover2016node2vec,FeatureNorm2020,feng2025mhr}, protein interaction networks~\cite{vazquez2003modeling,pogany2025hyperbolic,ren2025multifractal}, recommender systems~\cite{chen2021modeling,yang2022hrcf,yang2022hicf}, knowledge graphs~\cite{wang2017knowledge,lin2025multi,zheng2024low} (KGs), and financial transaction systems~\cite{sawhney2021exploring,caputi2024integral,fu2024hc}. 
Their widespread utilization enables the rapid storage and access of relational knowledge about interacting entities, forming the basis of numerous systems. 
Despite their expressive power, effectively extracting meaningful patterns from complex relational data remains a core challenge in machine learning. 
Graph-structured data is inherently rich in relational and topological information, but its irregular structure poses significant difficulties for traditional (Euclidean) machine learning models. 
\begin{figure}[t] 
\centering 
\includegraphics[width=0.3\textwidth]{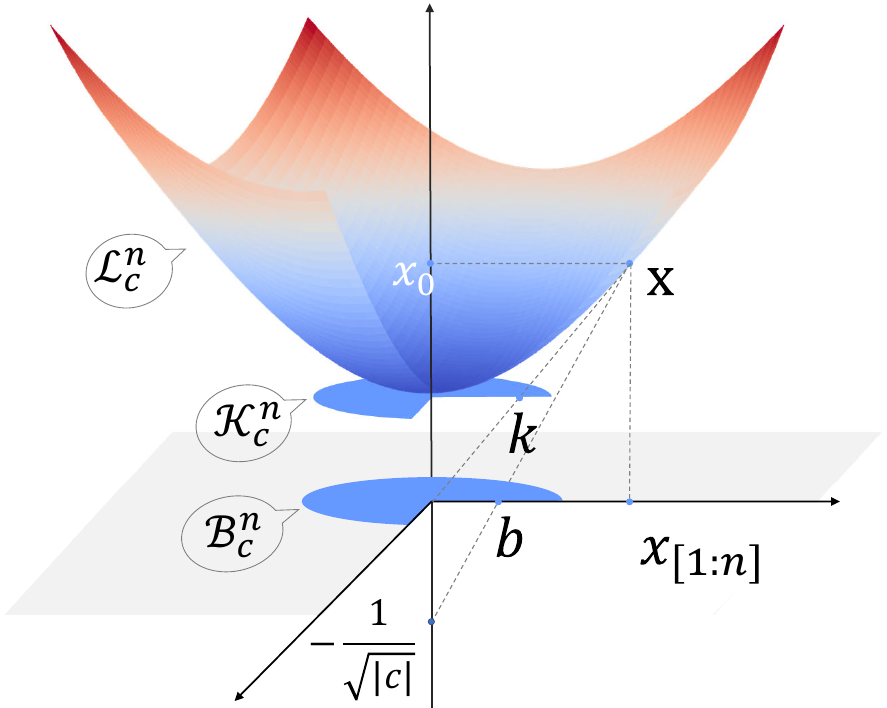}
\caption{Illustration of three prevalent and isomorphic hyperbolic graph models with their geometric relationships. The figure shows the Lorentz model $\mathcal{L}_c^n$ (hyperboloid in the upper portion), the Klein model $\mathcal{K}_c^n$ (disk in the middle), and the Poincaré ball model $\mathcal{B}_c^n$ (disk at the bottom) with curvature $c(c<0)$ or radius $1/\sqrt{|c|}$. The models are connected through bijective mappings, where a point $\mathbf{x}$ in the Lorentz model corresponds to the point $\mathbf{k}$ in the Klein model and the point $\mathbf{b}$ in the Poincar\'e ball model. 
}
\label{fig.three_manifolds}
\vspace{-10pt}
\end{figure}
\begin{figure*}[t] 
\centering 
\includegraphics[width=0.71\textwidth]{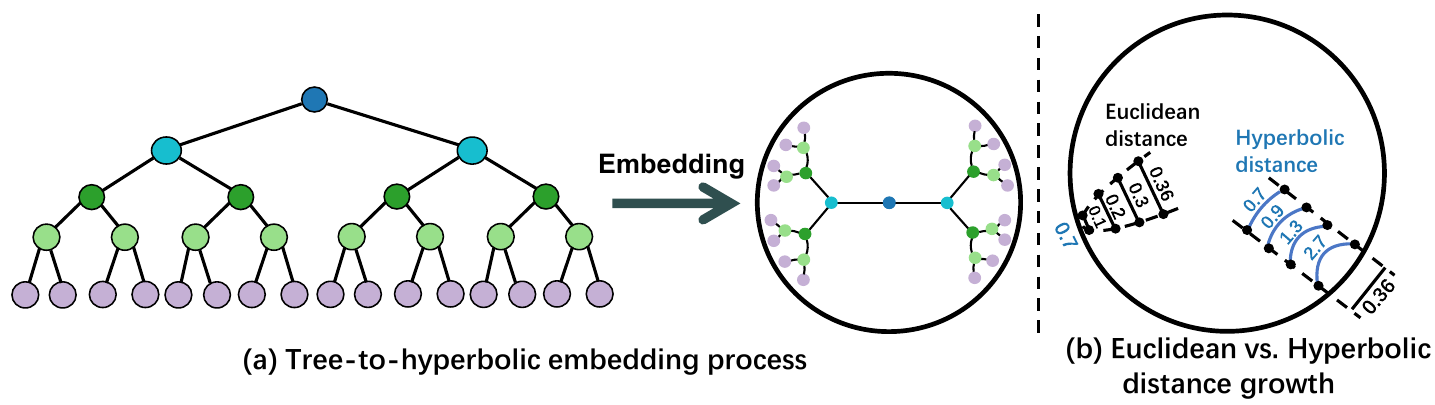}
\caption{Illustration of the geometric growth property and low-distortion embedding advantages of hyperbolic space. (a) Tree-to-hyperbolic embedding process: A hierarchical tree structure (left) is embedded into the Poincaré disk model (right), where nodes closer to the root are positioned near the center and leaf nodes are distributed toward the boundary, naturally preserving the tree's hierarchical relationships with minimal distortion. (b) Euclidean vs. Hyperbolic distance growth: Comparison of distance scaling in Euclidean space (left circle, showing linear distance intervals) versus hyperbolic space (right circle, showing exponentially increasing intervals toward the boundary). The hyperbolic space exhibits exponential volume growth with respect to radius, providing significantly more representational capacity than the polynomial growth in Euclidean space, making it particularly suitable for modeling hierarchical and scale-free graph structures.
}
\label{fig.Distortion_Figure}
\end{figure*}

To address these challenges, graph representation learning has emerged as a powerful paradigm, aiming to transform graph data into low-dimensional, informative embeddings that preserve structural properties and node semantics. Three prominent approaches in this area are \textit{graph embedding methods}, \textit{graph neural networks (GNNs)}, and \textit{emerging graph learning paradigms}. 
\textit{\textbf{Graph embedding} }methods~\cite{grover2016node2vec,goyal2018graph,cai2018comprehensive} focus on mapping nodes and edges into a continuous vector space, where proximity reflects similarity or structural relationships. 
\textit{\textbf{GNNs}} extend conventional neural networks to operate directly on graph-structured data through message-passing mechanisms. They enable learning of low-dimensional embeddings for nodes, edges, or entire graphs, which can be applied to downstream tasks such as node classification~\cite{gcn2017,velivckovic2017graph,sgc}, link prediction~\cite{kipf2016variational,zhang2018link}, and graph classification~\cite{GIN,errica2019fair}. 
\textit{\textbf{Emerging graph learning paradigms}} encompass advanced techniques such as diffusion-based graph generation and contrastive learning approaches that leverage novel learning objectives to capture complex graph structures.
These methods are traditionally built upon Euclidean space, which provides a vector space structure with closed-form expressions for distance and inner product, and aligns with our intuitive understanding of three-dimensional geometry~\cite{HNN}.

{\textbf{Why do we need hyperbolic space for graph Learning?}} Despite the significant achievements of Euclidean-based graph representation models, their performance remains constrained by the inherent representational capacity of Euclidean geometry (polynomially expanding capacity). Although nonlinear techniques~\cite{non_linear_embedding} assist in mitigating this issue, complex graph patterns may still need an embedding dimensionality that is computationally intractable. As revealed by Bronstein et al.~\cite{bronstein2017geometric}, many complex data show non-Euclidean underlying anatomy. Real-world graphs often exhibit complex structural properties that are poorly captured by flat Euclidean spaces. Specifically, many real-world graphs—such as social hierarchies, biological taxonomies, and KGs—exhibit tree-like structures, power-law degree distributions, and strong hierarchical organization.\footnote{The power-law distribution can be traced back to the hierarchical structures~\cite{ravasz2003hierarchical}.} Representing such structures efficiently in Euclidean space often requires high-dimensional embeddings, leading to computational inefficiency and potential distortion of intrinsic distances and relationships.

\textbf{Advantages of hyperbolic space.} Recently, hyperbolic geometry, characterized by its constant negative curvature, has offered a compelling alternative for embedding and learning from such non-Euclidean data. Fig.~\ref{fig.three_manifolds} illustrates three commonly used models of hyperbolic space, which are mathematically isomorphic. 
A key geometric property of hyperbolic space is its exponential volume growth with respect to radius, in contrast to the polynomial growth of Euclidean space. 
This property yields two main advantages, making hyperbolic space particularly suitable for representing tree-like graph structures. 
\textit{First}, hyperbolic space enables \textit{\textbf{low-distortion embedding}}, effectively preserving the original graph's topological structure when mapping to the embedding space. This is particularly beneficial for representing scale-free or hierarchical structured data, as illustrated in Fig.~\ref{fig.Distortion_Figure}(a). 
\textit{Secondly}, due to the \textit{\textbf{exponential distance scaling property}} of hyperbolic space (as shown in Fig.~\ref{fig.Distortion_Figure}(b)), 
the distance intervals between concentric circles increase exponentially toward the boundary, compared to the linear distance intervals in Euclidean space. This exponential scaling provides significantly more representational capacity than the polynomial growth in Euclidean space, enabling hyperbolic models to accommodate exponentially expanding hierarchical structures. 
As a result, hyperbolic models can capture complex structural information even in low-dimensional spaces, making them particularly suitable for resource-constrained scenarios such as low-memory or low-storage environments.

While the field of hyperbolic graph learning (HGL) is rapidly expanding, a comprehensive and unified review that systematically categorizes methods and thoroughly explores their diverse applications is still needed. 
Although existing surveys~\cite{peng2021hyperbolic,mettes2024hyperbolic} have reviewed hyperbolic learning from general and computer vision perspectives, respectively, they do not provide a dedicated and systematic analysis of HGL. In particular, fundamental challenges in HGL—such as preserving topological hierarchies in hyperbolic embeddings and enabling geometry-aware message passing—remain largely open and call for further investigation. 
To bridge this gap, this survey provides an in-depth and up-to-date overview of the current landscape of HGL, aiming to serve as a comprehensive resource for both researchers and practitioners in the field. 
\textbf{The main contributions} of this work are summarized below: 

\begin{itemize}
  \item \textbf{Comprehensive methodology taxonomy}: We present the first systematic categorization of HGL methods into three main paradigms: hyperbolic embedding techniques, GNN-based hyperbolic models, and emerging hyperbolic graph learning paradigms, providing a unified framework for understanding the field's evolution.
  
  \item \textbf{Extensive application survey}: We systematically examine HGL applications across four major domains—recommender systems, KGs, bioinformatics, and other emerging scenarios—demonstrating the broad applicability and effectiveness of hyperbolic geometry in real-world graph learning tasks.
  
  \item \textbf{Critical analysis and future roadmap}: We identify several key challenges in HGL and provide concrete research opportunities, establishing a foundation for future research directions in this rapidly evolving field.
\end{itemize}

\section{Preliminaries and Notation}
\label{sec2}
In this section, we briefly introduce a list of some of the most helpful concepts, definitions, and operations in hyperbolic geometry. For a more detailed introduction, please refer to~\cite{lee2013smooth} and \cite{JohnMLee1997RiemannianMA}.

\subsection{Mathematical Preliminaries}
\textbf{Manifold and Tangent Space}. Riemannian geometry is a subfield of differential geometry in which a smooth manifold $\mathcal{M}$ is associated with a Riemannian metric $g^{\mathcal{M}}$. An $n$-dimensional manifold $(\mathcal{M},g^\mathcal{M})$ is a topological space, a generalization of a 2-dimensional surface with high dimensions. 

\begin{table*}[t]
    \centering
        \caption{Summary of Operations in the Poincar{\'e} Ball Model, the Lorentz Model, and the Klein Model ($c<0$)}
        \resizebox{1.0\textwidth}{!}{%
            \begin{tabular}{l|c|c|c}
                \toprule
                & \quad\quad\quad\textbf{Poincar{\'e} Ball Model} & \quad\quad\textbf{Lorentz Model} & \quad\quad\textbf{Klein Model} \\
                \midrule
                \textbf{Manifold} & 
                $\mathcal{B}_{c}^{n}=\left\{\mathbf{x} \in \mathbb{R}^{n}:\langle \mathbf{x}, \mathbf{x}\rangle_{2}<-{1}/{c}\right\}$ & 
                $\mathcal{L}_{c}^{n}=\left\{\mathbf{x} \in \mathbb{R}^{n+1}:\langle \mathbf{x}, \mathbf{x}\rangle_{\mathcal{L}}={1}/{c}\right\}$ & 
                $\mathcal{K}_{c}^{n}=\left\{\mathbf{x} \in \mathbb{R}^{n}:\langle \mathbf{x}, \mathbf{x}\rangle_{2}<-{1}/{c}\right\}$ \\
                
                \textbf{Metric} & 
                $g_{\mathbf{x}}^{\mathcal{B}_{c}^n}=\left(\lambda_{\mathbf{x}}^{c}\right)^{2} g^{\mathrm{E}^n}$, where $\lambda_{\mathbf{x}}^{c}=\frac{2}{1+c\|\mathbf{x}\|_{2}^{2}}$ and $g^{\mathrm{E}^n}=\mathbf{I}_{n}$ & 
                $g_{\mathbf{x}}^{\mathcal{L}_{c}^{n}}=\frac{1}{|c|}\cdot\text{diag}(-1,1,\ldots,1)$ & 
                $g_{\mathbf{x}}^{\mathcal{K}_{c}^{n}}=\frac{1}{-c(1+c\|\mathbf{x}\|^2)}\left(\delta_{ij}+\frac{cx_ix_j}{1+c\|\mathbf{x}\|^2}\right)$, where $\delta_{ij}$ is the Kronecker delta \\
                
                \textbf{Induced distance} & 
                $d_{\mathcal{B}}^{c}(\mathbf{x}, \mathbf{y})=\frac{1}{\sqrt{|c|}} \cosh^{-1}\left(1-\frac{2 c\|\mathbf{x}-\mathbf{y}\|_{2}^{2}}{\left(1+c\|\mathbf{x}\|_{2}^{2}\right)\left(1+c\|\mathbf{y}\|_{2}^{2}\right)}\right)$ & 
                $d_{\mathcal{L}}^{c}(\mathbf{x}, \mathbf{y})=\frac{1}{\sqrt{|c|}} \cosh^{-1}\left(c\langle \mathbf{x}, \mathbf{y}\rangle_{\mathcal{L}}\right)$ & 
                $d_{\mathcal{K}}^{c}(\mathbf{x}, \mathbf{y})=\frac{1}{\sqrt{|c|}}\cosh^{-1}\left(\frac{1-|c|\langle\mathbf{x},\mathbf{y}\rangle}{\sqrt{(1-|c|\|\mathbf{x}\|^2)(1-|c|\|\mathbf{y}\|^2)}}\right)$ \\
                
                \textbf{Logarithmic map} & 
                $\log _{\mathbf{x}}^{c}(\mathbf{y})=\frac{2}{\sqrt{|c| \lambda_\mathbf{x}^{c}}} \tanh^{-1}\left(\sqrt{|c|}\left\|-\mathbf{x} \oplus_{c} \mathbf{y}\right\|_{2}\right) \frac{-\mathbf{x} \oplus_{c} \mathbf{y}}{\left\|-\mathbf{x} \oplus_{c} \mathbf{y}\right\|_{2}}$ & 
                $\log _{\mathbf{x}}^{c}(\mathbf{y})=\frac{\cosh^{-1}\left(c\langle \mathbf{x}, \mathbf{y}\rangle_{\mathcal{L}}\right)}{\sinh \left(\cosh^{-1}\left(c\langle \mathbf{x}, \mathbf{y}\rangle_{\mathcal{L}}\right)\right)}\left(\mathbf{y}-c\langle \mathbf{x}, \mathbf{y}\rangle_{\mathcal{L}} \mathbf{x}\right)$ & 
                $\log _{\mathbf{x}}^{c}(\mathbf{y})=\frac{d_{\mathcal{K}}^{c}(\mathbf{x},\mathbf{y})}{\|\mathbf{y}-\mathbf{x}\|_{\mathbf{x}}^{\mathcal{K}}} \cdot (\mathbf{y}-\mathbf{x})$ \\
                
                \textbf{Exponential map} & 
                $\exp _{\mathbf{x}}^{c}(\mathbf{v})=\mathbf{x} \oplus_{c}\left(\tanh \left(\sqrt{|c|} \frac{\lambda_{\mathbf{x}}^{c}\|\mathbf{v}\|_{2}}{2}\right) \frac{\mathbf{v}}{\sqrt{|c|}\|\mathbf{v}\|_{2}}\right)$ & 
                $\exp _{\mathbf{x}}^{c}(\mathbf{v})=\cosh \left(\sqrt{|c|}\|\mathbf{v}\|_{\mathcal{L}}\right) \mathbf{x}+\mathbf{v} \frac{\sinh \left(\sqrt{|c|}\|\mathbf{v}\|_{\mathcal{L}}\right)}{\sqrt{|c|}\|\mathbf{v}\|_{\mathcal{L}}}$ & 
                $\exp _{\mathbf{x}}^{c}(\mathbf{v})=\mathbf{x}+\frac{\tanh(\sqrt{|c|}\|\mathbf{v}\|_{\mathbf{x}}^{\mathcal{K}})}{\sqrt{|c|}\|\mathbf{v}\|_{\mathbf{x}}^{\mathcal{K}}} \mathbf{v}$ \\
                
                \textbf{Parallel transport} & 
                $PT_{\mathbf{x} \rightarrow \mathbf{y}}^{c}(\mathbf{v})=\frac{\lambda_{\mathbf{x}}^{c}}{\lambda_{\mathbf{y}}^{c}} \operatorname{gyr}[\mathbf{y},-\mathbf{x}] \mathbf{v}$ & 
                $PT_{\mathbf{x} \rightarrow \mathbf{y}}^{c}(\mathbf{v})=\mathbf{v}-\frac{c\langle \mathbf{y}, \mathbf{v}\rangle_{\mathcal{L}}}{1+c\langle \mathbf{x}, \mathbf{y}\rangle_{\mathcal{L}}}(\mathbf{x}+\mathbf{y})$ & 
                $PT_{\mathbf{x} \rightarrow \mathbf{y}}^{c}(\mathbf{v})=\mathbf{v}+\frac{c\langle\mathbf{y}-\mathbf{x},\mathbf{v}\rangle_{\mathbf{x}}^{\mathcal{K}}}{1+\sqrt{1+c\|\mathbf{x}\|^2}\sqrt{1+c\|\mathbf{y}\|^2}-c\langle\mathbf{x},\mathbf{y}\rangle}(\mathbf{y}-\mathbf{x})$ \\
                \bottomrule
            \end{tabular}
        }
    \begin{minipage}{\textwidth}
        \scriptsize
        \textbf{Note.} The Klein norm at point $\mathbf{x}$ is given by: 
        $\|\mathbf{v}\|_{\mathbf{x}}^{\mathcal{K}} = \sqrt{\frac{\|\mathbf{v}\|^2 - c(\langle\mathbf{x},\mathbf{v}\rangle)^2}{1+c\|\mathbf{x}\|^2}}$;
        The Klein inner product at point $\mathbf{x}$ is given by: 
        $\langle\mathbf{u}, \mathbf{v}\rangle_{\mathbf{x}}^{\mathcal{K}} = \frac{\langle\mathbf{u},\mathbf{v}\rangle - c\langle\mathbf{x},\mathbf{u}\rangle\langle\mathbf{x},\mathbf{v}\rangle}{1+c\|\mathbf{x}\|^2}$;
        All inner products $\langle \cdot, \cdot \rangle$ without a superscript refer to standard Euclidean inner products.
    \end{minipage}
    \label{tab:all_operations_on_poincare_hyperboloid}
    \vspace{-5pt}
\end{table*}

For each point $\mathbf{x}$ in $\mathcal{M}$, a tangent space $\mathcal{T}_\mathbf{x}\mathcal{M}$ is defined as the first-order approximation of $\mathcal{M}$ around $\mathbf{x}$, which is an $n$-dimension vector space and isomorphic to $\mathbb{R}^n$. The Riemannian manifold metric $g^\mathcal{M}$ assigns a smoothly varying positive definite inner product $<\cdot,\cdot>:\mathcal{T}_\mathbf{x}\mathcal{M}\times \mathcal{T}_\mathbf{x}\mathcal{M}\to \mathbb{R}$ on the tangent space, which allows us to define several geometric properties, such as geodesic distances, angles, and curvature. 

 \textbf{Geodesics and Induced Distance Function}. For a curve $\gamma:[\alpha,\beta]\to \mathcal{M}$, the shortest length of $\gamma$, i.e., geodesics, is defined as $L(\gamma)=\int_\alpha^\beta\|\gamma^\prime(t)\|_g dt$. Then the distance of $\mathbf{u}, \mathbf{v} \in \mathcal{M}$, $d_\mathcal{M}(\mathbf{u},\mathbf{v})=\inf L(\gamma)$ where $\gamma$ is a curve that $\gamma(\alpha)=\mathbf{u}, \gamma(\beta)=\mathbf{v}$. 

\begin{figure}[t] 
\centering 
\includegraphics[width=0.3\textwidth]{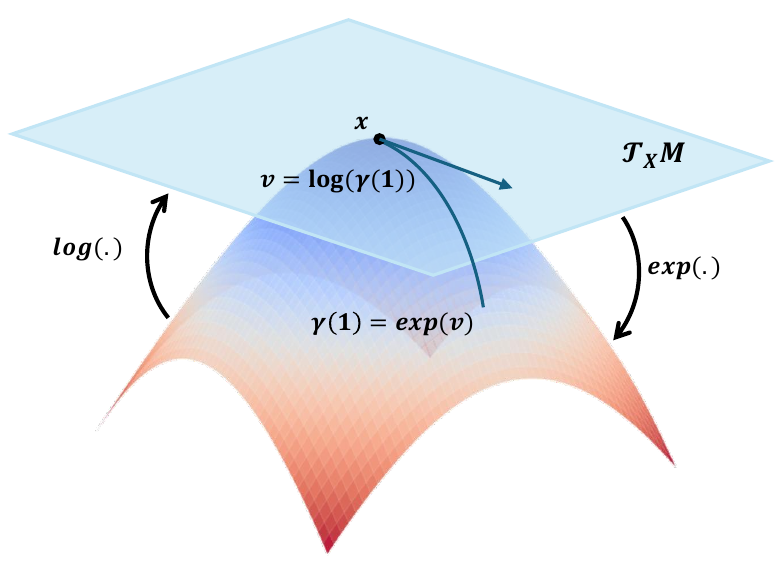}
\caption{Illustration of exponential and logarithmic maps between a hyperbolic manifold $\mathcal{M}$ and the tangent space $\mathcal{T}_\mathbf{x}\mathcal{M}$ at point $\mathbf{x}$. A tangent vector $\mathbf{v}\in\mathcal{T}_\mathbf{x}\mathcal{M}$ is mapped onto the manifold via the exponential map, producing a point $\gamma(1)=\exp(\mathbf{v})\in\mathcal{M}$. Conversely, a point on the manifold can be projected back to the tangent space using the logarithmic map, $\log(\gamma(1))=\mathbf{v}$.
}
\label{fig.2.1}
\vspace{-10pt}
\end{figure}
    
\textbf{Maps and Parallel Transport}. 
The maps define the relationship between the hyperbolic space and the corresponding tangent space. As shown in Fig.~\ref{fig.2.1}, for a point $\mathbf{x}\in \mathcal{M}$ and vector $\mathbf{v}\in\mathcal{T}_\mathbf{x}\mathcal{M}$, there exists a unique geodesic $\gamma:[0,1]\to\mathcal{M}$ where $\gamma(0)=\mathbf{x}, \gamma^\prime(0)=\mathbf{v}$. The exponential map $\exp_\mathbf{x}: \mathcal{T}_\mathbf{x}\mathcal{M} \to \mathcal{M}$ is defined as $\exp_\mathbf{x}(\mathbf{v})=\gamma(1)$ and logarithmic map $\log_\mathbf{x}$ is the inverse of $\exp_\mathbf{x}$. The parallel transport $PT_{\mathbf{x}\rightarrow \mathbf{y}}:\mathcal{T}_\mathbf{x}\mathcal{M}\to\mathcal{T}_\mathbf{y}\mathcal{M}$ achieves the transportation from point $\mathbf{x}$ to $\mathbf{y}$ that preserves the metric tensors.

\subsection{Isometric Models in the Hyperbolic Space}
Hyperbolic geometry is a Riemannian manifold with a constant negative curvature. As shown in Fig.~\ref{fig.three_manifolds}, there exist multiple equivalent hyperbolic models, like the Poincar\'e ball model, Lorentz model, and Klein model, which show different characteristics but are mathematically equivalent. 
The three models, commonly used in HGL, are given by Definition~\ref{def:poincare}, Definition~\ref{def:lorentz} and 
Definition~\ref{def:klein}, respectively. 
We summarize related formulas and operations for the Poincar\'e ball Model, the Lorentz model and the Klein model in Table~\ref{tab:all_operations_on_poincare_hyperboloid}, where  $\oplus_{c}$ and $\operatorname{gyr}[.,.] v$ are M\"obius addition~\cite{ungar2007hyperbolic} and gyration operator~\cite{ungar2007hyperbolic}, respectively. 
Finally, we presented the bijective formula for the relationship between spaces as defined in Definition~\ref{def:bijections}. 
   
\begin{definition}[Poincar\'e Ball Model]
With negative curvature $c\quad(c<0)$, the Poincar\'e ball model is defined as a Riemannian manifold $(\mathcal{B}_{c}^{n},g_{\mathbf{x}}^{\mathcal{B}})$, where $\mathcal{B}_{c}^{n}=\left \{\mathbf{x}\in \mathbb{R}^{n}:\| \mathbf{x} \|^2 < -1/c \right \}$ is an open $n$-dimensional ball with radius $1/\sqrt{|c|}$. Its metric tensor $g_{\mathbf{x}}^{\mathcal{B}}=(\lambda_\mathbf{x}^{c})^2g^{E}$, known as the Poincar\'e metric, is defined such that points closer to the boundary of the ball are "further away" in hyperbolic distance, where $\lambda_\mathbf{x}^c=2/({1+c\|\mathbf{x}\|_2^2})$ is the conformal factor and $g^{E}$ is the Euclidean metric, i.e., $\mathbf{I}_n$. 
\label{def:poincare}
\end{definition}
        
\begin{definition}[Lorentz Model] With negative curvature $c\quad(c<0)$, the Lorentz model (also named hyperboloid model) defines $n$-dimensional hyperbolic space as the upper sheet of a two-sheeted hyperboloid embedded in $(n+1)$-dimensional Minkowski space. 
Minkowski space is equipped with the Minkowski inner product $\left \langle \mathbf{x},\mathbf{y} \right \rangle_{\mathcal{L}}=-x_0y_0+\sum_{i=1}^{n} x_iy_i$. The hyperboloid is defined as the Riemannian manifold $( \mathcal{L}_{c}^{n}, g_\mathbf{x}^{\mathcal{L}_{c}^{n}})$, where $\mathcal{L}_{c}^{n} = \left \{  \mathbf{x}\in \mathbb{R}^{n+1}: \left \langle \mathbf{x},\mathbf{x} \right \rangle_{\mathcal{L}}={1}/{c} \right \}$ and its metric tensor $g_{\mathbf{x}}^{\mathcal{L}_{c}^{n}}=\frac{1}{|c|}\cdot\text{diag}([-1,1,...,1])_{n}$. 
\label{def:lorentz}
\end{definition}

\begin{definition}[Klein Model]
With negative curvature $c\quad(c<0)$, the Klein model, also known as the projective model, represents $n$-dimensional hyperbolic space as the open unit ball in $\mathbb{R}^{n}$, similar to the Poincar\'e ball. The Klein model is obtained by mapping $\mathbf{x}\in \mathcal{L}_{c}^{n}$ to the hyperplane $x_0=1/\sqrt{|c|}$, using rays emanating from the origin. Formally, the Klein model is defined as $\mathcal{K}_{c}^{n}=\left \{\mathbf{x}\in \mathbb{R}^{n}:\| \mathbf{x} \|^2 < -1/c \right \}$. However, unlike the Poincar\'e ball, geodesics in the Klein model are represented by straight line segments. 
\label{def:klein}
\end{definition}

\begin{definition}[Isometric Isomorphism]
The Lorentz model $\mathcal{L}$, the Poincar\'e ball model $\mathcal{B}$ and the Klein model $\mathcal{K}$ are geometrically isomorphic. With negative curvature $c\quad(c<0)$, the bijections between a node $\mathbf{x} = [x_0,x_1,...,x_n]\in\mathcal{L}_{c}^{n}$ and its unique corresponding node $\mathbf{b} = [b_0,b_1,...,b_{n-1}]\in\mathcal{B}_{c}^{n}$ are given by
\begin{equation}
p_{\mathcal{L}_{c}^{n}\to\mathcal{B}_{c}^{n}}(\mathbf{x})=[\frac{x_1}{x_0+\sqrt{\frac{1}{|c|}}},\cdots,\frac{x_n}{x_0+\sqrt{\frac{1}{|c|}}}],
\end{equation}
\begin{equation}
p_{\mathcal{B}_{c}^{n}\to\mathcal{L}_{c}^{n}}(\mathbf{b})=[\frac{1+|c|\|\mathbf{b}\|^2}{\sqrt{|c|}(1-|c|\|\mathbf{b}\|^2)},\frac{2\mathbf{b}}{1-|c|\|\mathbf{b}\|^2}].
\end{equation}
The bijections $p$ between $\mathbf{x} = [x_0,x_1,...,x_n]\in\mathcal{L}_{c}^{n}$ and its unique corresponding node $\mathbf{k} = [k_0,k_1,...,k_{n-1}]\in\mathcal{K}_{c}^{n}$ are given by
\begin{equation}
p_{\mathcal{L}_{c}^{n}\to\mathcal{K}_{c}^{n}}(\mathbf{x})=[\frac{x_1}{x_0},\cdots,\frac{x_n}{x_0}],
\end{equation}
\begin{equation}
p_{\mathcal{K}_{c}^{n}\to\mathcal{L}_{c}^{n}}(\mathbf{k})=[\frac{1}{\sqrt{|c|(1-|c|\|\mathbf{k}\|^2)}},\frac{\mathbf{k}}{\sqrt{1-|c|\|\mathbf{k}\|^2}}].
\end{equation}
\label{def:bijections}
\end{definition}

\textbf{Key Properties and Distinctions.} Each hyperbolic model offers unique computational and geometric advantages that make them suitable for different applications. 
The \textit{\textbf{Poincaré ball model}} exhibits conformality, meaning that angles between intersecting curves are preserved from the Euclidean representation. This property makes it visually intuitive for understanding hyperbolic geometry and simplifies certain geometric operations, particularly those involving angular relationships and rotations.
The \textit{\textbf{Lorentz model}} is often considered the most "natural" or intrinsic representation of hyperbolic space, as its geometry directly arises from the Minkowski inner product. This intrinsic nature simplifies the definition of fundamental geometric operations such as parallel transport and exponential maps, making it computationally advantageous for theoretical analysis and algorithmic implementations.
The \textit{\textbf{Klein model}} simplifies geodesics to straight line segments, facilitating efficient distance computations and path planning. However, it is not conformal, meaning that angles are generally distorted compared to their Euclidean counterparts. This model is particularly useful in projective geometry and for understanding the global structure of hyperbolic space due to its straight-line geodesic property.

In the following sections, we use $\mathcal{H}$ to represent cases that are applicable to all three models $\mathcal{B}$, $\mathcal{L}$, and $\mathcal{K}$. However, when specific geometric properties or computational advantages are relevant, we employ more precise notation to describe the particular model being utilized.

\section{Hyperbolic Graph Representation Methods}
\label{sec3}
The unique geometric properties of hyperbolic space, particularly its exponential volume growth, make it inherently suitable for modeling complex hierarchical and tree-like structures prevalent in real-world graphs. This section provides a comprehensive overview of the methodologies developed for learning graph representations in hyperbolic spaces. We begin by discussing the \textit{Hyperbolic Initialization (\textsection\ref{sec: initial_layer})}, a crucial first step for many hyperbolic methods. Subsequently, we categorize the core approaches into \textit{\textbf{Hyperbolic Graph Embedding} (\textsection\ref{sec: Hyperbolic Embedding})} techniques and \textit{\textbf{Hyperbolic GNN-based Models} (\textsection\ref{sec: GNN-based Hyperbolic Models}}), along with \textit{\textbf{Emerging Hyperbolic Graph Learning Paradigms} (\textsection\ref{sec: Other Hyperbolic Graph Learning Paradigms})}, like diffusion-based and contrastive learning approaches. Fig.~\ref{fig: classification of HGL} provides an overview of some key methods and related models discussed in this section.

\tikzstyle{leaf}=[draw=hiddendraw,
    rounded corners,minimum height=1em,
    fill=mygreen!40,text opacity=1, align=center,
    fill opacity=.5,  text=black,align=left,font=\scriptsize,
    inner xsep=3pt,
    inner ysep=1pt,
    ]
\tikzstyle{middle}=[draw=hiddendraw,
    rounded corners,minimum height=1em,
    fill=output-white!40,text opacity=1, align=center,
    fill opacity=.5,  text=black,align=left,font=\scriptsize,
    inner xsep=3pt,
    inner ysep=1pt,
    ]
\begin{figure*}[ht]
\centering
\begin{forest}
  for tree={
  forked edges,
  grow=east,
  reversed=true,
  anchor=base west,
  parent anchor=east,
  child anchor=west,
  base=middle,
  font=\scriptsize,
  rectangle,
  line width=0.7pt,
  draw=output-black,
  rounded corners,align=left,
  minimum width=2em,
    s sep=5pt,
    inner xsep=3pt,
    inner ysep=1pt,
  },
  where level=1{text width=4.5em}{},
  where level=2{text width=6em,font=\scriptsize}{},
  where level=3{font=\scriptsize}{},
  where level=4{font=\scriptsize}{},
  where level=5{font=\scriptsize}{},
  [Hyperbolic Graph Representation, middle,rotate=90,anchor=north,edge=output-black
    [Hyperbolic Initialization \\\quad \quad\quad\quad\quad(\textsection\ref{sec: initial_layer}), middle, edge=output-black, text width=8.7em
        [Pre-trained Embedding (\textsection\ref{sec:Pre-trained Embedding}), middle, text width=12.0em, edge=output-black
            [
            BERT-H~\cite{chen2021probing}{,}
            HyperQA~\cite{tay2018hyperbolic}{,}
            HGCN~\cite{hgcn2019}{,}
            HGNN~\cite{liu2019HGNN}{,}
            HGAT~\cite{zhang2021hyperbolic}
            , leaf, text width=22.3em, edge=output-black]
        ]
        [Random Sampling (\textsection\ref{sec:Random Sampling}), middle, text width=12.0em, edge=output-black
            [
            Poincar\'eEmb~\cite{nickel2017poincare}{,}
            HGCH~\cite{zhang2024hgch}{,}
            HGCF~\cite{sun2021hgcf}{,}
            HRCF~\cite{yang2022hrcf}{,}
            HICF~\cite{yang2022hicf}{,}
            \\LorentzEmb~\cite{nickel2018learning}{,}
            HVGNN~\cite{sun2021hyperbolic}
            , leaf, text width=22.3em, edge=output-black]
        ]
        [One-hot Encoding (\textsection\ref{sec:One-hot Encoding}), middle, text width=12.0em, edge=output-black
            [
            Qu-HNN~\cite{qu2022hyperbolic}{,}
            HTGN~\cite{yang2022hyperbolic_htgn,yang2021discrete}
            , leaf, text width=22.3em, edge=output-black]
        ]
    ]
    [\quad \quad Hyperbolic Graph \\\quad\quad Embedding (\textsection\ref{sec: Hyperbolic Embedding}), middle, edge=output-black, text width=8.7em
        [Distance-based Methods (\textsection\ref{sec: Distance-based Embeddings}) , middle, text width=12.0em, edge=output-black
            [
            Poincar\'eEmb~\cite{nickel2017poincare}{,}
            LorentzEmb~\cite{nickel2018learning}{,}
            LorentzDL~\cite{law2019lorentzian}{,}
            FPS-T~\cite{cho2023curve}{,}
            \\M$^2$GNN~\cite{wang2021mixed}{,}
            Gu-MuRP~\cite{gu2019learning}
            , leaf, text width=22.3em, edge=output-black]
        ]
        [Angle-based Methods (\textsection\ref{sec: Angle-based Methods}), middle, text width=12.0em, edge=output-black
            [
            HECone~\cite{ganea2018hyperbolic}{,}
            ConE~\cite{bai2021modeling}{,}
            HSCone~\cite{yu2023shadow}
            , leaf, text width=22.3em, edge=output-black]
        ]
        [Disk-based Methods (\textsection\ref{sec: Disk-based Methods}), middle, text width=12.0em, edge=output-black
            [
            HDiskEmb~\cite{suzuki2019hyperbolic}, leaf, text width=22.3em, edge=output-black]
        ]
    ]
    [Hyperbolic GNN models\\ \quad \quad \quad\quad\quad(\textsection\ref{sec: GNN-based Hyperbolic Models}), middle, edge=output-black, text width=8.7em
        [Tangent Space-based Models (\textsection\ref{sec:Tangent Space-based Models}), middle, text width=12.0em, edge=output-black
            [HGNN~\cite{liu2019HGNN}{,}
            HyperKA~\cite{sun2020knowledge}{,}
            H$^2$E~\cite{wang2021knowledge}{,}
            HGCN~\cite{hgcn2019}{,}
            LGCN~\cite{lgcn}
            ,leaf, text width=22.3em, edge=output-black]
        ]
        [Fully Hyperbolic Models (\textsection\ref{sec:Fully Hyperbolic Models}), middle, text width=12.0em, edge=output-black
            [HYBONET~\cite{chen2021fully}{,}
            H2H-GCN~\cite{dai2021hyperbolic}{,}
            LGCF~\cite{wang2021fully}{,}
            FFHR~\cite{shi2023ffhr}
            ,leaf, text width= 22.3em, edge=output-black]
        ]
        [\quad Neighborhood Aggregation \\  \quad\quad\quad Mechanism (\textsection\ref{sec:weight_computation}), middle, text width=12.0em, edge=output-black
            [
            HGNN~\cite{liu2019HGNN}{,}
        $\kappa$HGCN~\cite{yang2023kappahgcn}{,}
        HGCN~\cite{hgcn2019}{,}
        HAT~\cite{zhang2021hyperbolic}{,}
        LGCN~\cite{lgcn}{,}
        \\HYBONET~\cite{chen2021fully}{,}
        HAN~\cite{HAT}{,}
        Conehead~\cite{tseng2023coneheads}{,}
        GIL~\cite{zhu2020graph}
        ,leaf, text width= 22.3em, edge=output-black]
        ]
    ]
    [\quad Emerging Hyperbolic \\ \quad Graph Learning (\textsection\ref{sec: Other Hyperbolic Graph Learning Paradigms}), middle, edge=output-black, text width=8.7em
        [Diffusion-based Methods (\textsection\ref{sec:Diffusion-based Hyperbolic Representations}), middle, text width=12.0em, edge=output-black
            [HypDiff~\cite{fu2024hyperbolic}{,}
            HGDM~\cite{wen2024hyperbolic}{,}
            HDRM~\cite{yuan2025hyperbolic}
            , leaf, text width=22.3em, edge=output-black]
        ]
        [Graph Contrastive Learning (\textsection\ref{sec: Hyperbolic Graph Contrastive Learning}), middle, text width=12.0em, edge=output-black
            [HGCL~\cite{liu2022enhancing}{,}
            HyperGCL~\cite{zhang2023alignment}  
            , leaf, text width=22.3em, edge=output-black]
        ]
    ]
  ]
\end{forest}
\caption{A comprehensive taxonomy of Hyperbolic Graph Learning (HGL) methods organized into four main categories: Hyperbolic Initialization (§3.1), Hyperbolic Graph Embedding (§3.2), Hyperbolic GNN models (§3.3), and Emerging Hyperbolic Graph Learning (§3.4). Each branch shows the subcategories and lists representative papers with reference numbers, providing a structured overview of the HGL methodological landscape.}
\label{fig: classification of HGL}
\vspace{-10pt}
\end{figure*}

\subsection{Hyperbolic Initialization}
\label{sec: initial_layer}
Consider a graph $\mathcal{G}=(\mathcal{V}, \mathcal{E})$ with vertex set $\mathcal{V}$ and edge set $\mathcal{E}$. For each vertex $i \in \mathcal{V}$, its $n$-dimensional input node feature is denoted as $\mathbf{x}_i^{E}$, residing in Euclidean space, where superscript $^E$ indicates the Euclidean space. To refer to the tangent state at $\mathbf{x}$ and the hyperbolic space, we use superscripts $^{\mathcal{T}_\mathbf{x}}$ and $^{\mathcal{H}(\mathcal{B}/\mathcal{L}/\mathcal{K})}$, respectively.

Hyperbolic embeddings can be obtained through two main processes: direct initialization in hyperbolic space and then projection from Euclidean space via exponential maps.
Euclidean node features $\mathbf{x}^E$ are typically derived from three primary sources: pre-trained embeddings, random sampling strategies, and one-hot encoded representations.

\textbf{Pre-trained Embedding.}\label{sec:Pre-trained Embedding}
This approach initializes node features using embeddings pre-trained on various tasks~\cite{tay2018hyperbolic,chen2021probing}, which capture semantic meanings through contextual, neighborhood, or structural information~\cite{hgcn2019,liu2019HGNN,zhang2021hyperbolic}. Node features can also be derived directly from inherent node attributes when available, enabling models to leverage explicit characteristics alongside structural information. Recent advances in large language models have further enhanced this approach—methods like HyperLLM~\cite{cheng2025large} extract semantic hierarchies from textual descriptions, creating initial embeddings with parent-child structures that naturally align with hyperbolic geometry's inductive biases.

\textbf{Random Sampling.}\label{sec:Random Sampling}
When pre-trained features or node attributes are unavailable, random initialization provides a flexible alternative. Standard approaches sample from uniform or Gaussian distributions~\cite{nickel2017poincare,nickel2018learning,sun2021hgcf,yang2022hrcf,yang2022hicf,sun2021hyperbolic}, offering simplicity without requiring prior knowledge. More sophisticated variants leverage domain-specific priors—HGCH~\cite{zhang2024hgch} exploits the Poincaré space's conformal property to propose power-law initialization for recommender systems:
\begin{equation}
\mathbf{x}_i^{E}\sim\mathrm{Uni}(-an_i^{-b},an_i^{-b}),
\label{equ:power-law prior distributions}
\end{equation}
where $n_i$ denotes node frequency, $a$ controls the initialization scale, and $b$ parameterizes the power-law distribution to better capture hierarchical structures.

\textbf{One-hot Encoding.}\label{sec:One-hot Encoding}
In scenarios requiring explicit node identity preservation, each node is represented by a unique one-hot encoded vector with dimensionality equal to the graph size~\cite{qu2022hyperbolic,yang2022hyperbolic_htgn,yang2021discrete}. While highly sparse, this representation ensures complete node distinguishability and is particularly useful when learning embeddings from scratch without external features.

After obtaining Euclidean node features $\mathbf{x}^E$ through any of the above initialization strategies, they must be mapped to hyperbolic space to leverage its geometric properties. The exponential map provides the standard mechanism for this projection, with implementation details varying across hyperbolic models.

For the \textit{Poincaré ball model}, the projection is straightforward:
\begin{equation}
    \mathbf{x}^{\mathcal{B}} = \exp_\mathbf{o}^c(\mathbf{x}^E),
    \label{equ:init_poincare_ball}
\end{equation}
where $\mathbf{x}^E$ is interpreted as a vector in the tangent space at the origin, i.e., $\mathbf{x}^E\in \mathcal{T}_\mathbf{o}\mathcal{B}^n$.

The \textit{Lorentz model} requires additional consideration due to its embedding in Minkowski space. Since tangent vectors at the origin must satisfy orthogonality constraints, $\mathbf{x}^{E}$ is augmented with a zero time component~\cite{hgcn2019}: $\mathbf{x}^{\mathcal{T}_\mathbf{o}} = (0, \mathbf{x}^{E})$, ensuring $\langle 0, \mathbf{x}^{\mathcal{T}_\mathbf{o}} \rangle_\mathcal{L} = 0$. The projection then follows:
\begin{equation}
\mathbf{x}^{\mathcal{L}}=\exp_\mathbf{o}^c(\mathbf{x}^{\mathcal{T}_\mathbf{o}})=\exp_\mathbf{o}^c((0,\mathbf{x}^E)).
\label{equ:init_lorentz}
\end{equation}
Alternative approaches include pseudo polar coordinate projections~\cite{HAT}, which yield equivalent results to Equation~(\ref{equ:init_lorentz}).

\textbf{Discussion and Summary.} This initialization ensures that node representations are geometrically consistent within hyperbolic space, providing a solid foundation for subsequent layers to learn and propagate information while preserving the beneficial properties of hyperbolic geometry.

\subsection{Hyperbolic Graph Embedding}
\label{sec: Hyperbolic Embedding}
\begin{figure*}[t] 
\centering 
\includegraphics[width=0.7\textwidth]{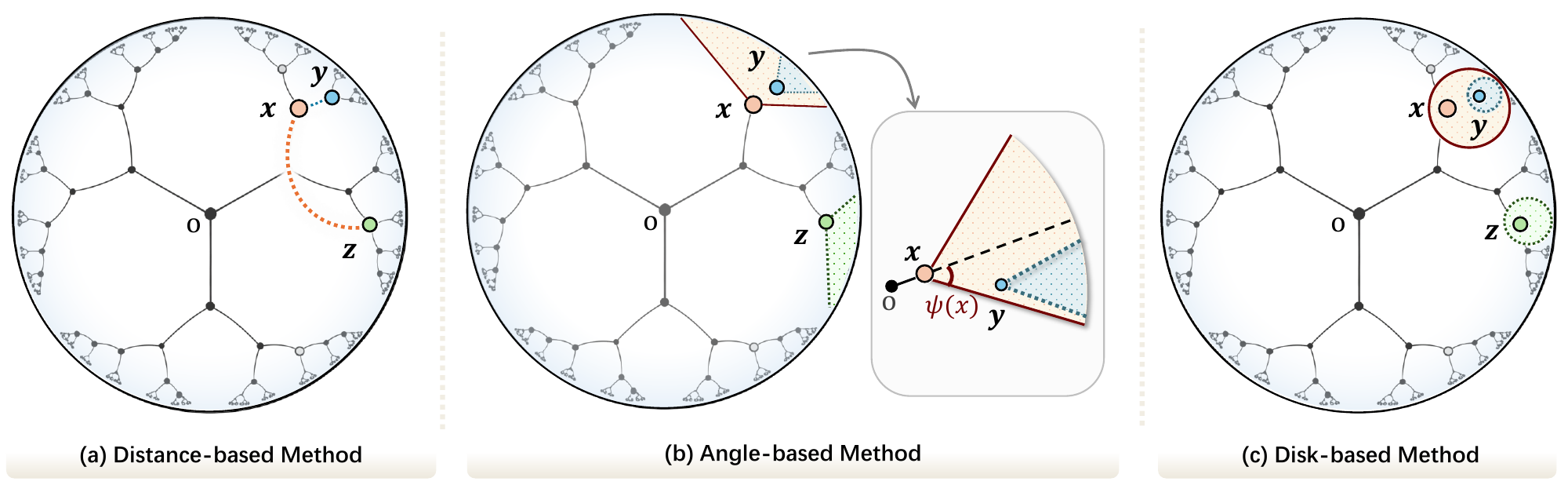}
\vspace{-10pt}
\caption{
Illustration of three categories of hyperbolic representation methods. In each subfigure, $x$ and $y$ form a positive sample pair with a hierarchical relationship (e.g., parent–child), while $x$ and $z$ are negative pairs. (a) Distance-based methods preserve hierarchy by placing $x$ and $y$ closer in terms of hyperbolic distance, while pushing $z$ away. (b) Angle-based methods encode partial order (e.g., subset inclusion) via entailment cones: the cone rooted at $x$ (the apex) encloses the one at $y$, indicating that $x$ is more general. (c) Disk-based methods use region-based embeddings to model containment or uncertainty, where the disk of $x$ contains that of $y$.
}
\label{fig.embedding}
\vspace{-10pt}
\end{figure*}

Hyperbolic embedding methods represent a foundational approach to HGL. Unlike GNN-based models that emphasize feature propagation within hyperbolic manifolds, these methods focus on directly mapping graph elements—such as nodes, edges, or entire subgraphs—into hyperbolic space, where geometric relations (e.g., distances, angles) are designed to preserve the structural and semantic properties of the original graph. 
This section delves into various approaches for hyperbolic embedding, as shown in Fig.~\ref{fig.embedding}. We categorize existing hyperbolic embedding approaches by the primary geometric properties: 
\textit{Distance-based Methods (\textsection\ref{sec: Distance-based Embeddings})}, which use hyperbolic distance to encode hierarchical structures or relational proximity; 
\textit{Angle-based Methods (\textsection\ref{sec: Angle-based Methods})}, which model partial order relations (e.g., hierarchical dependencies or subset inclusion) via entailment cones; 
\textit{Disk-based Methods~(\textsection\ref{sec: Disk-based Methods})}, which employ region-based representations to capture containment relationships or model uncertainty. 

\subsubsection{Distance-based Methods}
\label{sec: Distance-based Embeddings}
Distance-based hyperbolic embeddings leverage hyperbolic distance metrics to encode graph structure, positioning nodes such that their geometric proximity reflects relational similarity in the original graph. This paradigm naturally captures hierarchical relationships by implicitly embedding root nodes near the hyperbolic space center and placing descendant nodes toward the boundary, leveraging the exponential expansion of hyperbolic space~\cite{nickel2017poincare,nickel2018learning}.

A seminal work in this area is Poincar\'eEmb~\cite{nickel2017poincare}, which demonstrated the Poincar\'e ball model's effectiveness for tree-like symbolic data:
\begin{equation}
\mathbf{\ell}_\text{pdist}=\sum_{(\mathbf{x},\mathbf{y})\in\mathcal{D}}\log\frac{e^{-d^{c}_{\mathcal{B}}(\mathbf{x}^{\mathcal{B}},\mathbf{y}^{\mathcal{B}})}}{\sum_{\mathbf{y}^{\prime}\in\mathcal{N}(\mathbf{x})}e^{-d^{c}_{\mathcal{B}}(\mathbf{x}^{\mathcal{B}},\mathbf{y}^{\prime \mathcal{B}})}},
\end{equation}
This loss function $\mathbf{\ell}_{pdist}$ employs a softmax formulation over hyperbolic distances, where $\mathcal{D}$ represents positive pairs and $\mathcal{N}(\mathbf{x})$ denotes negative samples for node $\mathbf{x}$. The objective encourages similar nodes to have smaller hyperbolic distances while pushing dissimilar nodes apart. This approach laid the foundation for hyperbolic representation learning by achieving orders-of-magnitude improvements over Euclidean methods on hierarchical datasets such as WordNet.

To address numerical instabilities inherent in Poincar\'e distance computations, alternative formulations emerged using the Lorentz model. LorentzEmb~\cite{nickel2018learning} explored the hyperboloid representation using standard Lorentz distances, while LorentzDL~\cite{law2019lorentzian} specifically focused on squared Lorentz distances to enable closed-form centroid calculations:
\begin{equation}
\mathbf{\ell}_\text{ldist}=\sum_{(\mathbf{x},\mathbf{y})\in\mathcal{D}}\log\frac{e^{-(d^{c}_{\mathcal{L}}(\mathbf{x}^{\mathcal{L}},\mathbf{y}^{\mathcal{L}}))^2}}{\sum_{\mathbf{y}^{\prime}\in\mathcal{N}(x)}e^{-(d^{c}_{\mathcal{L}}(\mathbf{x}^{\mathcal{L}},\mathbf{y}^{\prime \mathcal{L}}))^2}},
\end{equation}
where the squared Lorentzian distance $(\mathsf{d}_{\mathcal{L}}^{c}(\mathbf{x}^{\mathcal{L}},\mathbf{y}^{\mathcal{L}}))^2=\frac{2}{c}-2\langle \mathbf{x}^{\mathcal{L}},\mathbf{y}^{\mathcal{L}}\rangle_{\mathcal{L}}$ in LorentzDL enables efficient centroid computation. A key insight from LorentzDL is that centroid Euclidean norms decrease with space curvature, naturally encoding hierarchical structures where parent nodes possess smaller norms than their children.

 Early distance-based methods assume fixed-curvature spaces, but real-world graphs often exhibit heterogeneous structures that cannot be faithfully captured by single-curvature geometries. To address this limitation, Gu-MuRP~\cite{gu2019learning} introduces product manifolds combining hyperbolic, Euclidean, and spherical spaces, enabling task-specific curvature allocation. This framework has been extended to KG completion~\cite{wang2021mixed} and attention mechanisms~\cite{cho2023curve}, demonstrating superior performance by accommodating diverse graph topologies through multiple geometric inductive biases.

\subsubsection{Angle-based Methods}
\label{sec: Angle-based Methods}
While distance-based methods in hyperbolic space have shown strong capability in modeling hierarchical structures, they often struggle to represent partial orders and entailment relations directly. To address this, a class of angle-based methods~\cite{ganea2018hyperbolic,bai2021modeling,yu2023shadow,du2022hakg}. 
Introduces hyperbolic cones to model hierarchy via inclusion relations, leveraging the angular structure of hyperbolic space to encode soft entailments and partial orders more explicitly. These methods typically leverage the geometry of hyperbolic cones or regions to represent the "scope" or "reach" of an entity, thereby naturally encoding transitivity and asymmetry.

HECone~\cite{ganea2018hyperbolic} treats hierarchical relationships as partial orders defined by a family of nested geodesic convex cones within the Poincar\'e model. 
Specifically, in a vector space, a convex cone $S$ (with apex at the origin) is defined as a set that is closed under non-negative linear combinations. That is, for any two vectors $v_1, v_2 \in S$ and any non-negative real numbers $\alpha, \beta \geq 0$, the linear combination $\alpha v_1 + \beta v_2$ also belongs to $S$. 
The authors proposed to build the hyperbolic cones using the exponential map, taking any cone in the tangent space $S\subseteq \mathcal{T}_{x}\mathcal{M}$ at a fixed point $x$ and map it into a set $\mathfrak{S}_{x}\subset\mathcal{M}$:
\begin{equation}
\mathfrak{S}_{x}:=\exp_{x}^{c}\left(S\right),\quad S\subseteq \mathcal{T}_{x}\mathcal{M}.
\end{equation}
To avoid heavy cone intersections and ensure capacity that scales exponentially with the dimensionality of the space, angular cones are constructed within the Poincar\'e ball. The angular cones meet four intuitive properties, including axial symmetry, rotation invariance, continuity of cone aperture function, and the transitivity of nested angular cones. 
HECone~\cite{ganea2018hyperbolic} further provided a closed-form of the angular cones, for any $\mathbf{x},\mathbf{y}\in\mathcal{B}^n\setminus\mathcal{B}^n(O,\epsilon)$, the angle between the half-lines $({x}{y}$ and $({O}{x}$ as:
\begin{equation}
\Xi(x,y):=\pi-\angle Oxy.
\end{equation}
Then, this angle equals:
\begin{equation}
\arccos(\frac{\langle\mathbf{x},\mathbf{y}\rangle(1+\left\|\mathbf{x}\right\|^{2})-\left\|\mathbf{x}\right\|^{2}(1+\left\|\mathbf{y}\right\|^{2})}{\left\|\mathbf{x}\right\|\left\|\mathbf{x}-\mathbf{y}\right\|\sqrt{1+\left\|\mathbf{x}\right\|^{2}\left\|\mathbf{y}\right\|^{2}-2\left\langle\mathbf{x},\mathbf{y}\right\rangle}}).
\end{equation}
Then, the equivalent expression of the Poincaré entailment cones is
\begin{equation}
\left.\mathfrak{S}_{x}^{\psi(x)}=\left\{
\begin{array}
{cc}y\in\mathcal{B}^n & \Xi(x,y)\leq\arcsin\left(K\frac{1-\|x\|^2}{\|x\|}\right)
\end{array}\right.\right\},
\end{equation}
where $\psi(x)=\arcsin\left(K\frac{1-\|x\|^2}{\|x\|}\right)$ is cone aperture functions, $K$ is a constant. The corresponding loss function can be defined as:
\begin{equation}
E(x,y):=\max(0,\Xi(x,y)-\psi(x)),
\end{equation}
\begin{equation}
\mathbf{\ell}_\text{HECone}=\sum_{(x,y)\in\mathcal{T}}E(x,y)+\sum_{(x,y)\in\mathcal{N}}\max(0,\mu-E(x,y)),
\end{equation}
where $\mathcal{T}$ and $\mathcal{N}$ are sets of positive and negative samples, $\mu$ is a constant margin.

To address the complexities of graphs with heterogeneous structure, ConE~\cite{bai2021modeling} represents the embedding space of the KG triplet $(h,r,t)$ as the product space of $d$ two-dimensional hyperbolic planes. For entity 
$h$, $\textbf{h}=(\textbf{h}_1,...,\textbf{h}_d)$ where $\textbf{h}_i\in\mathcal{B}^{2}$ is the apex of the $i$-th 2D hyperbolic cone. 
ConE adopts rotation transformation $f_1$ to model non-hierarchical properties and $f_2$ to model hierarchical properties. 
\begin{equation}
f_1(\mathbf{h}_i,\mathbf{r}_i)=\exp^c_{\mathbf{o}}(\mathbf{G}(\theta_i)\log_{\mathbf{o}}(\mathbf{h}_i)),
\end{equation}
\begin{equation}
f_2(\mathbf{h}_i,\mathbf{r}_i)=\exp^c_{\mathbf{h}_i}(s_i\cdot\mathbf{G}(\theta_i\frac{\psi(\mathbf{h}_i)}{\pi})\overline{\mathbf{h}}_i),\mathbf{r}_i=(s_i,\theta_i),
\end{equation}
where $s_i,\theta_i$ corresponds to scaling and rotation; $\mathbf{G}(\theta_i)$ is the Givens rotation matrix: 
\begin{equation}
\mathbf{G}(\theta_i)=
\begin{bmatrix}
\cos(\theta_i) & -\sin(\theta_i) \\
\sin(\theta_i) & \cos(\theta_i)
\end{bmatrix};
\end{equation}
$\overline{\mathbf{h}}_i$ is the unit vector of $\mathbf{h}_i$ in the tangent space of $\mathbf{h}_i$:
\begin{equation}
\overline{\mathbf{h}}_i=\widehat{\mathbf{h}}_i/||\widehat{\mathbf{h}}_i||,\widehat{\mathbf{h}}_i=\log_{\mathbf{h}_i}(\frac{1+||\mathbf{h}_i||}{2||\mathbf{h}_i||}\mathbf{h}_i),
\end{equation}
On this basis, the angle loss is introduced (without loss of generality let $r$ be a hyponym relation):
\begin{equation}
\mathbf{\ell}_\text{ConE}=\mathbf{m}_{\mathbf{r}_i}\cdot(\max(0,\Xi(\mathbf{h}_i,\mathbf{t}_i)-\psi(\mathbf{h}_i)))_{i\in\{1,\cdots,d\}},
\end{equation}
the subspace can be partitioned using a hyperbolic plane mask $\mathbf{m}$, where each element $\mathbf{m}_i \in \{0, 1\}$. A value of $\mathbf{m}_i = 1$ indicates that cone containment is enforced in the $i$-th hyperbolic plane. 
This design enables ConE to impose localized constraints within specific subspaces of the embedding space via the angular cone formulation, facilitating the effective use of hyperbolic cones in modeling heterogeneous graphs. 

Further generalizing the framework of partial order embeddings, HSCone~\cite{yu2023shadow} proposes the "Shadow Cone"—a physics-inspired entailment structure. Drawing inspiration from the shadows cast by celestial bodies, it models partial orders as shadow cones formed by the obstruction of light from a source. 
To adapt to different data types and capture varying degrees of entailment, they propose specific types of shadow cones, namely the Umbral cone and Penumbral cone. This framework offers a more flexible and robust way to embed partial orders by defining cones based on a novel projection mechanism. This method allows for greater flexibility in capturing the intricate relationships within partial orders, potentially reducing distortion and improving the accuracy of entailment prediction compared to simpler cone-based approaches.

\subsubsection{Disk-based Methods}
\label{sec: Disk-based Methods}
While traditional hyperbolic embedding methods often represent nodes as single points or angle-based cones, disk-based embeddings~\cite{tsuiki2008lawson,goubault2013few} extend the representation of a node from a singular point to a hyperbolic disk or region. 
This can be viewed as a higher-dimensional representation compared to one-dimensional distance-based methods (point-to-point relationships) or two-dimensional angle-based cone methods (cone-based regions). This richer representation allows for capturing more complex topological structures and relationships.

Existing embedding methods, particularly those designed for tree-like hierarchies, often struggle with complex graphs where the number of both ancestors and descendants can grow exponentially. 
Such scenarios are common in DAGs, which are more general than trees and pose significant challenges for distortion-free embedding. 
HDiskEmb~\cite{suzuki2019hyperbolic} extends the disk embedding framework to hyperbolic geometry, specifically utilizing the Poincar\'e disk model. The core innovation lies in its ability to represent DAGs by embedding each node as a hyperbolic disk, where disk containment naturally encodes reachability or partial-order relations. Let $C=\{c_i\}_{i=1}^N$ denote a set of entities with partial order relations $\preceq$, and let $(X,d)$ be a quasi-metric space. Disk embeddings are defined as a mapping $c_i\mapsto(\mathbf{x}_i,r_i)$ such that $c_i \preceq c_j$ if and only if $(\mathbf{x}_i,r_i)\sqsubseteq(\mathbf{x}_j,r_j)$, where the latter denotes disk containment. The protrusion parameter between two formal disks $(\mathbf{x}_i,r_i)$ and $(\mathbf{x}_j,r_j)$ is $l_{ij}=l(\mathbf{x}_i,r_i;\mathbf{x}_j,r_j)=d(\mathbf{x}_i,\mathbf{x}_j)-r_i+r_j$. The corresponding loss function can be defined as:
\begin{equation}
\mathbf{\ell}_\text{disk}=\sum_{(i,j)\in\mathcal{T}}h_+(l_{ij})+\sum_{(i,j)\in\mathcal{N}}h_+(\mu-l_{ij}),
\end{equation}
where $\mathcal{T}$ and $\mathcal{N}$ are sets of positive and negative samples, $\mu$ is a constant margin, $h_+(x) = max(0, x)$. This framework generalizes both entailment cones and order embeddings, as they can be regarded as special cases of hyperbolic disk embeddings. 

\textbf{Discussion and Summary.} Distance-based methods~\cite{nickel2017poincare,nickel2018learning,law2019lorentzian} and angle-based methods~\cite{ganea2018hyperbolic,bai2021modeling,yu2023shadow} extend the embedding constraints from one-dimensional to two-dimensional structures, while disk-based methods~\cite{suzuki2019hyperbolic} further generalize the embeddings themselves to higher-dimensional representations. These advancements enable more effective utilization of the expressive capacity of hyperbolic space.

\subsection{GNN-based Hyperbolic Models}
\label{sec: GNN-based Hyperbolic Models}
\begin{figure*}[t] 
\centering 
\includegraphics[width=0.95\textwidth]{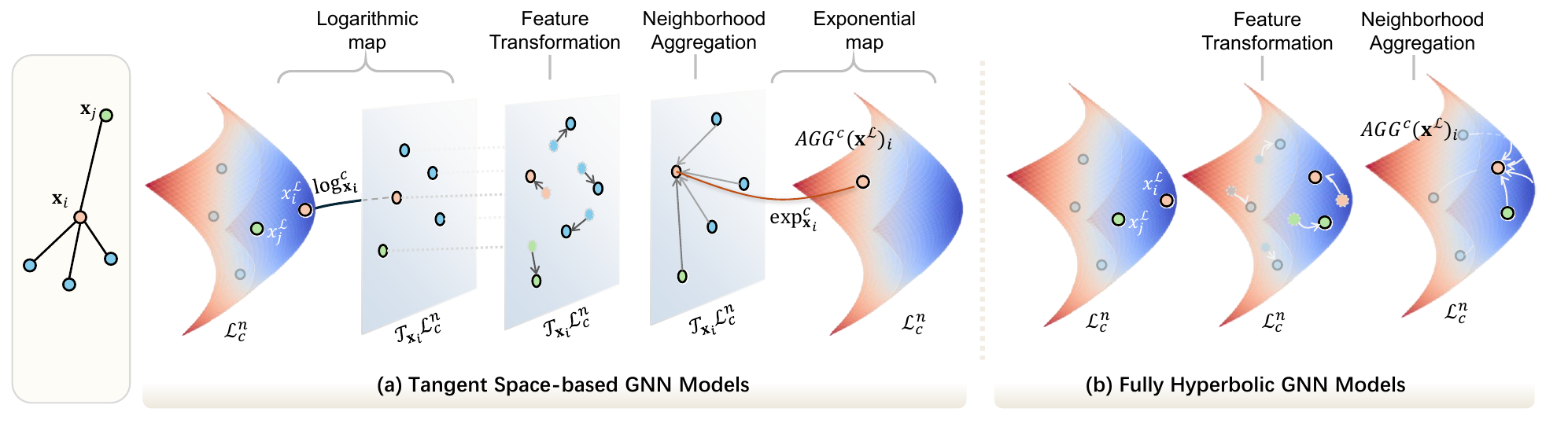}
\vspace{-10pt}
\caption{Comparison between (a) tangent space-based and (b) fully hyperbolic GNN models. In (a), node features are first mapped from the hyperbolic space to the tangent space via logarithmic mapping for Euclidean-style feature transformation and aggregation, and then projected back using the exponential map. In contrast, (b) performs both feature transformation and neighborhood aggregation directly within the hyperbolic manifold, preserving the space's geometric inductive bias throughout the process.}
\label{fig.gnn}
\vspace{-10pt}
\end{figure*}

While hyperbolic embedding methods~\cite{nickel2017poincare,ganea2018hyperbolic,suzuki2019hyperbolic} directly map graph entities into hyperbolic spaces, 
Hyperbolic Graph Neural Networks (HGNNs) integrate the message-passing paradigm with non-Euclidean manifolds~\cite{liu2019HGNN,yang2023kappahgcn,hgcn2019,zhang2021hyperbolic}. These models aim to take advantage of the superior capacity of hyperbolic spaces for hierarchical data while benefiting from the powerful feature learning and aggregation capabilities of GNNs. Broadly, as shown in Fig.~\ref{fig.gnn}, HGNNs can be categorized into two main paradigms based on how they handle computations within the curved manifold: \textit{Tangent Space-based Models (\textsection\ref{sec:Tangent Space-based Models})} and \textit{Fully Hyperbolic Models~(\textsection\ref{sec:Fully Hyperbolic Models})}. Each approach presents distinct advantages and limitations. 

\subsubsection{Tangent Space-based Models}
\label{sec:Tangent Space-based Models}

As shown in Fig.~\ref{fig.gnn}(a), tangent space-based models typically project hyperbolic embeddings to a local Euclidean tangent space, perform operations in this flat space, and then project the results back to the hyperbolic manifold. Following the typical GNN learning paradigm, tangent space-based models implement three fundamental operations in the hyperbolic setting: \textit{hyperbolic feature transformation, hyperbolic neighborhood aggregation}, and \textit{non-linear activation}. 

\textbf{Hyperbolic Feature Transformation.} In tangent space-based models, feature transformations (mainly consisting of matrix-vector multiplication and bias addition) are typically performed after mapping node embeddings from the hyperbolic manifold to a tangent space. Specifically, for node embedding in Poincar\'e ball model $\mathbf{x}^{\mathcal{B}}\in \mathcal{B}^{n}_c, \mathbf{W}\in \mathbb{R}^{d\times n}$, the matrix-vector multiplication~\cite{HNN,liu2019HGNN} is defined by
\begin{equation}
\label{equ:poincare_matrix_multiplication}
        \mathbf{W} \otimes_{c}^\mathcal{B} \mathbf{x}^\mathcal{B}:= \exp_\mathbf{o}^c(\mathbf{W}\log_\mathbf{o}^c(\mathbf{x}^\mathcal{B})).
\end{equation}
The node features 
$\mathbf{x}^\mathcal{B}$ are projected from the Poincar\'e ball model to the tangent space, which is isometric to Euclidean space. After performing the matrix-vector multiplication operation in the tangent space, they are mapped back to the Poincar\'e ball.

In the context of the Lorentz model, one method to achieve matrix-vector multiplication is 
similar to that of the Poincar\'e ball model. Chami et al.~\cite{hgcn2019} directly employed Equation~(\ref{equ:poincare_matrix_multiplication}) on the Lorentz model. However, this approach breaks the constraints inherent to the Lorentz manifold. To rectify this, it is imperative that the node features remain within the tangent space at the origin after the matrix $\mathbf{W}$ multiplication\footnote{Supposing we use Lorentz origin as the reference point.}. As suggested by \cite{lgcn,yang2022hyperbolic_htgn}, this can be accomplished by modifying the values of the last $n$ coordinates. The formula is defined by
\begin{equation}
\label{equ:Lorentz_matrix_multiplication}
        \mathbf{W} \otimes^{\mathcal{L}}_{c} \mathbf{x}^\mathcal{L}:= \exp_\mathbf{o}^c(0, \mathbf{W}\log_\mathbf{o}^c(\mathbf{x}^\mathcal{L})_{[1:n]}),
\end{equation}
where $\mathbf{x}^{\mathcal{L}}\in \mathcal{L}^{n}_c, \mathbf{W}\in \mathbb{R}^{d\times n}$.
This method ensures the first coordinate is consistently zero, signifying that the resultant transformation is invariably within the tangent space at $\mathbf{o}$. 

To implement bias addition, the tangent space at the origin is still a useful medium, and the formula in $\mathcal{B}$ and $\mathcal{L}$ can be uniformly expressed as:
\begin{equation}
\mathbf{x}^\mathcal{H} \oplus_{c}^{\mathcal{H}} \mathbf{b}^\mathcal{H}=\exp_{\mathbf{x}^\mathcal{H}}^{c}\left(P_{\mathbf{o} \rightarrow \mathbf{x}^\mathcal{H}}^{c}\left(\log _{\mathbf{o}}^{c}(\mathbf{b}^\mathcal{H})\right)\right),
\end{equation}
where $\mathbf{b}^\mathcal{H}$ is the bias in $\mathcal{H}_c^n$, and the equations of parallel transport $P_{\mathbf{x}^\mathcal{H} \rightarrow \mathbf{y}^\mathcal{H}}^{c}(\cdot)$ in $\mathcal{B}$ and $\mathcal{L}$ are summarized in Table~\ref{tab:all_operations_on_poincare_hyperboloid}.

\textbf{Hyperbolic Neighborhood Aggregation.} Hyperbolic neighborhood aggregation is a technique that leverages the hyperbolic space to aggregate information from neighboring nodes in a graph. This process can be broadly divided into two main steps: computation of neighborhood weights and computation of mean aggregation. While both tangent space-based and fully hyperbolic models differ in how they transform and aggregate features, the strategy for computing neighborhood weights is conceptually orthogonal to whether the model operates in tangent or curved space. We present the neighborhood aggregation mechanism in Section~(\ref{sec:weight_computation}). For mean aggregation, or weighted mean pooling, it cannot be computed by simply averaging the inputs, which may lead to a deviation from the hyperbolic manifold. The tangential mean computation is one of the most straightforward methods. It is applicable to the Poincar\'e and Lorentz models. Tangential method~\cite{liu2019HGNN,hgcn2019} is defined by,
\begin{equation}
\mathrm{AGG}(\mathbf{x}^\mathcal{H}_i):=\exp_{\mathbf{o}}^{c}\left(\sum_{j\in\mathcal{N}_i} \alpha_{ij}\left(\log _{\mathbf{o}}^{c}(\mathbf{x}^\mathcal{H}_i)\right)\right).
\end{equation}
where $\alpha_{ij}$ represents the weight from the neighboring node. 

\textbf{Non-linear Activation.} The non-linear activation can be achieved with the same idea of matrix-vector multiplication~\cite{hgcn2019}, i.e.,
\begin{equation}
\sigma^{{c_{\mathbf{\ell}-1}, c_{\mathbf{\ell}}}}\left(\mathbf{x}^{\mathcal{H}}\right)=\exp _{\mathbf{o}}^{c_{\mathbf{\ell}}}\left(\sigma(\log _{\mathbf{o}}^{c_{\mathbf{\ell}-1}}(\mathbf{x}^{\mathcal{H}}))\right),
\end{equation}
where $\mathbf{\ell}$ denotes the $\mathbf{\ell}-th$ layer.
For the Lorentz model, \cite{lgcn} proposed a more accurate form, accomplished the calculation by using space-like features
\begin{equation}
\sigma^{{c_{\mathbf{\ell}-1}, c_{\mathbf{\ell}}}}\left(\mathbf{x}^{\mathcal{H}}\right)=\exp _{\mathbf{o}}^{c_{\mathbf{\ell}}}\left(0, \sigma(\log _{\mathbf{o}}^{c_{\mathbf{\ell}-1}}(\mathbf{x}^{\mathcal{H}}_{[1:n]}))\right),
\end{equation}
which ensures that the result still lives in the Lorentz manifold.

\subsubsection{Fully Hyperbolic Models}
\label{sec:Fully Hyperbolic Models}

Fully hyperbolic models~\cite{chen2021fully,dai2021hyperbolic,wang2021fully}, in contrast, aim to conduct most GNN operations directly on the hyperbolic manifold, avoiding repeated projections to the tangent space. 
Similarly, once we have obtained the initialization state of the nodes, we proceed with \textit{hyperbolic feature transformation} and \textit{hyperbolic neighborhood aggregation}, followed by \textit{non-linear activation}. 

\textbf{Hyperbolic Feature Transformation.} For Poinca\'e ball models, it's a good choice to use the Gyrovector space~\cite{ungar2001hyperbolic}, which is a generalization of Euclidean vector spaces to models of hyperbolic space based on M\"obius transformations. Specifically, for node embedding in Poincar\'e ball model, the M\"obius vector multiplication $\mathbf{W}{\otimes^\mathcal{B}}\mathbf{x}^\mathcal{B}$ between the matrix $\mathbf{W}$ and input $\mathbf{x}^\mathcal{B}$ is defined by:
\begin{equation}
\mathbf{W}{\otimes^\mathcal{B}}\mathbf{x}^\mathcal{B}=\tanh\left(\frac{||\mathbf{W}\mathbf{x}^\mathcal{B}||}{||\mathbf{x}^\mathcal{B}||}\mathrm{actanh}(||\mathbf{x}^\mathcal{B}||)\right)\frac{\mathbf{W}\mathbf{x}^\mathcal{B}}{||\mathbf{W}\mathbf{x}^\mathcal{B}||},
\end{equation}
The bias addition can be implement by M\"obius addition. In the Gyrovector space, the M\"obius addition $\oplus^{\mathcal{B}}$ for $\mathbf{x}^\mathcal{B}$ and $\mathbf{y}^\mathcal{B}$ in Poinca\'e ball is defined as
\begin{equation}
\mathbf{x}^\mathcal{B}\oplus^{\mathcal{B}} \mathbf{y}^\mathcal{B}=\frac{(1+2\langle \mathbf{x}^\mathcal{B},\mathbf{y}^\mathcal{B}\rangle+||\mathbf{y}^\mathcal{B}||^2)\mathbf{x}^\mathcal{B}+(1-||\mathbf{x}^\mathcal{B}||^2)\mathbf{y}^\mathcal{B}}{1+2\langle \mathbf{x}^\mathcal{B},\mathbf{y}^\mathcal{B}\rangle+||\mathbf{x}^\mathcal{B}||^2||\mathbf{y}^\mathcal{B}||^2}.
\end{equation}

For the Lorentz models, Dai et al.~\cite{dai2021hyperbolic} defined a new matrix feature transformation as: 
\begin{equation}
\begin{gathered}
\mathbf{W}\otimes^{\mathcal{L}}\mathbf{x}^{\mathcal{L}}:=\mathbf{W} \mathbf{x}^{\mathcal{L}} \\
\text { s.t. } \mathbf{W}=\left[\begin{array}{cc}
1 & \mathbf{0}^{\top} \\
\mathbf{0} & \widehat{\mathbf{M}}
\end{array}\right], \widehat{\mathbf{M}}^{\top} \widehat{\mathbf{M}}=\mathbf{I},
\end{gathered}
\end{equation}
where $\mathbf{W}\in \mathbb{R}^{n+1}$, $\widehat{\mathbf{M}}$ is a transformation sub-matrix, acting as a rotation matrix. $\mathbf{0}$ is a column vector of zeros, and $\mathbf{I}$ is an identity matrix.
Compared to Equation (\ref{equ:Lorentz_matrix_multiplication}), this method employs $\widehat{\mathbf{M}}$ to define $\mathbf{W}$. This transformation respects the Lorentz constraints, let $\mathbf{y}$ be the result, then we have the following:
\begin{equation}
\begin{aligned}
& \langle\mathbf{y}, \mathbf{y}\rangle_{\mathcal{L}}=-x_0^2+\left(\widehat{\mathbf{M}} \mathbf{x}_{1: n}\right)^{\top} \widehat{\mathbf{M}} \mathbf{x}_{1: n} \\
& =-x_0^2+\mathbf{x}_{1: n}^{\top}\left(\widehat{\mathbf{M}}^{\top} \widehat{\mathbf{M}}\right) \mathbf{x}_{1: n} \\
& =-x_0^2+\mathbf{x}_{1: n}^{\top} \mathbf{x}_{1: n} \\
& =-1.
\end{aligned} 
\end{equation}
Then, it is derived that the transformed result $\mathbf{y}$ lies in the Lorentz model. Recent study~\cite{chen2021fully} 
shows that the above linear transformation only considers a special rotation but no boost. They then
derived a general transformation method, which is equipped with both rotation and boost operations as well bias addition, normalization, etc., i.e.,
\begin{equation}
\mathbf{W}\otimes^\mathcal{L}_c\mathbf{x}^\mathcal{L}=\left(\left[\begin{array}{l}
\mathbf{v}^{\top} \\
\mathbf{M}
\end{array}
\right]\right)\otimes^\mathcal{L}_c\mathbf{x}^\mathcal{L}=\left[\begin{array}{c}
\sqrt{\|\phi(\mathbf{M x}, \mathbf{v})\|^2-1 / K} \\
\phi(\mathbf{M x}, \mathbf{v})
\label{equ:fully_lorentz}
\end{array}\right],
\end{equation}
where $\phi(\mathbf{M} \mathbf{x}, \mathbf{v})=$ $\frac{\lambda \sigma\left(\mathbf{v}^{\top} \mathbf{x}+b^{\prime}\right)}{\|\mathbf{M} h(\mathbf{x})+\mathbf{b}\|}(\mathbf{M} h(\mathbf{x})+\mathbf{b})$, where $\sigma$ is the sigmoid function, $\mathbf{b}$ and $b^{\prime}$ are bias terms, $\lambda>0$ controls the scaling range, $h$ is the activation function. In \cite{chen2021fully}, they incorporated the bias addition in Equation~(\ref{equ:fully_lorentz}).

\textbf{Hyperbolic Neighborhood Aggregation.} Just as mentioned by the tangent space-based method, we present the neighborhood aggregation mechanism in Section~(\ref{sec:weight_computation}). For the fully hyperbolic method, although the Fr\'echet mean operation~\cite{lou2020Frechet} lacks differentiability, there are three typical ways to implement mean aggregation: Einstein midpoint, Lorentzian centroid and M\"obius gyromidpoint: 

(1) \textit{Einstein midpoint} method~\cite{HAT,dai2021hyperbolic} is based on the Klein coordinates and applicable to Poincar\'e ball and Lorentz models by the isomorphic bijections:
\begin{equation}
\mathrm{AGG}(\mathbf{x}^\mathcal{H}_i):=\left\{\begin{array}{l}
\overline{\mathbf{x}}_{j}^{\mathcal{K}}=p_{\mathcal{H} \rightarrow \mathcal{K}}\left({\mathbf{x}}_{j}^{\mathcal{H}}\right) \\
\mathbf{m}_{i}^{\mathcal{K}}=\sum_{j \in {\mathcal{N}}_i} \gamma_{j} \overline{\mathbf{x}}_{j}^{\mathcal{K}} / \sum_{j \in {\mathcal{N}}_i} \gamma_{j}, \\
\mathbf{x}_{i}^{\mathcal{H}}=p_{\mathcal{K} \rightarrow \mathcal{H}}\left(\mathbf{m}_{i}^{\mathcal{K}}\right)
\end{array}\right.
\end{equation}
where $p_{\mathcal{M}_1 \rightarrow \mathcal{M}_2}$ denotes the projection from $\mathcal{M}_1$ to $\mathcal{M}_2$ and $
\gamma_{j}=(1-\|\overline{\mathbf{x}}_{j}^{\mathcal{K}}\|^{2})^{-1/2}
$ is the Lorentz factor.

(2) \textit{Lorentzian centroid} method~\cite{lgcn,chen2021fully,qu2022hyperbolic} is is designed for the Lorentz model and can be expressed as:
\begin{equation}
    \mathrm{AGG}(\mathbf{x}^\mathcal{L}_i):= \frac{\sum_{j\in\mathcal{N}_i}\alpha_{ij}\mathbf{x}^\mathcal{L}_j}{\sqrt{c}|\|\sum_{j\in\mathcal{N}_i}\alpha_{ij}\mathbf{x}^\mathcal{L}_j\|_\mathcal{L}|},
\end{equation}
where $\alpha_{ij}$ represents the weight from the neighboring node. 

(3) \textit{M\"obius gyromidpoint} method~\cite{ungar2008gyrovector} is ca losed-form expression for Poincar\'e model to compute the average (midpoint) in the Gyrovector spaces:
\begin{equation}
\mathrm{AGG}(\mathbf{x}^\mathcal{B}_i;\mathbf{\alpha})=\frac{1}{2}\oplus\left(\sum_{i=1}^N\frac{\alpha_i\gamma_i}{\sum_{j=1}^N\alpha_j(\gamma_j-1)}\mathbf{x}^\mathcal{B}_i\right)
\end{equation}
where $\mathbf{\alpha}=(\alpha_1,...,\alpha_N)$ as the weights for each sample $\mathbf{x}^\mathcal{B}_i$ and $\gamma_i=\frac{2}{||\mathbf{x}^\mathcal{B}_i||^2}$. 
These three centroids can be characterized as a minimizer of the weighted sum of calibrated squared distance~\cite{HNN++}.

\textbf{Non-linear Activation.} According to the manifold-preserving properties between the Lorentz model and Poincar\'e ball model, we can convert the Lorentzian feature to the Poincar\'e feature and implement the non-linearity in Poincar\'e ball model~\cite{dai2021hyperbolic}:
\begin{equation}
\sigma^{{c_{\mathbf{\ell}-1}, c_{\mathbf{\ell}}}}\left(\mathbf{x}^{\mathcal{H}}\right)=p_{\mathcal{B} \rightarrow \mathcal{H}}\left(\sigma\left(p_{\mathcal{H} \rightarrow \mathcal{B}}\left(\mathbf{x}^{{H}}\right)\right)\right)
\end{equation}
Chen et al.~\cite{chen2021fully} incorporated the non-lienar activation in Equation~(\ref{equ:fully_lorentz}).
Since the exponential map is a non-linear operation, which can produce the non-linearity, works by~\cite{sun2021hgcf,wang2021fully} thus ignored it.

\begin{table*}[!t]
    \centering
    \caption{Summary of neighborhood aggregation mechanism in HGNNs. SI: structure information; FI: feature information; HD: hyperbolic distance; HC: hyperbolic cone.}
    \resizebox{\textwidth}{!}{%
    \begin{tabular}{l|c|c|c}
        \toprule
        \textbf{Formula} & \textbf{Source} & \textbf{Reference} & \textbf{Description} \\
        \midrule
        \midrule
        $\alpha_{ij} = 1/\sqrt{\tilde{d}_i \tilde{d}_j}$ & SI & \cite{liu2019HGNN}& Simplified spectral convolution based on node degree $d_i,d_j$ \\
        $\alpha_{ij} = \text{softmax}_{j \in N(i)} \left(\text{MLP}(\kappa_{i,j}) \right)$ & SI & \cite{yang2023kappahgcn} & Curvature-based weight $\kappa$ sensitive to local graph structure \\
        $\alpha_{ij} = \mathrm{softmax}_{j \in N(i)}(\mathrm{LeakReLU}(\mathbf{W}^T [\mathbf{x}_i^{\mathcal{T}_\mathbf{o}}; \mathbf{x}_j^{\mathcal{T}_\mathbf{o}}]))$ & FI & \cite{hgcn2019}  &  Weighting adaptive to node feature $\mathbf{x}$ \\
        $\alpha_{ij} = \mathrm{softmax}_{j \in N(i)}(-d_c(\mathbf{x}_i^\mathcal{H}, \mathbf{x}_j^\mathcal{H}))$ & HD & \cite{zhang2021hyperbolic} &  Based on negative hyperbolic distance $d_c$ between nodes \\
        $\alpha_{ij} = \mathrm{softmax}_{j \in N(i)}(-d_c^2(\mathbf{x}_i^\mathcal{H}, \mathbf{x}_j^\mathcal{H}))$ & HD & \cite{lgcn,chen2021fully} & Weighting with squared hyperbolic distance, emphasizing geometry \\
        $\alpha_{ij} = f(-\beta d_c(\mathbf{x}_i^\mathcal{H}, \mathbf{x}_j^\mathcal{H}) - \gamma)$ & HD & \cite{HAT} & Flexible weighting incorporating squared hyperbolic distance $d_c^2$  \\
        $\alpha_{ij} = f(d_\mathcal{H}^c(\sup_2(\mathbf{x}_i^\mathcal{H},\mathbf{y}_i^\mathcal{H}),\mathcal{S}))$ & HC & \cite{tseng2023coneheads} & Weighting with the height of the hyperbolic cone  \\
        $\alpha_{ij} = \mathrm{softmax}_{j \in N(i)} (\mathrm{LeakReLU}(\mathbf{W}^T [\mathbf{x}_i; \mathbf{x}_j]) \times d_c(\mathbf{x}_i^\mathcal{H}, \mathbf{x}_j^\mathcal{H}))$ & FI,HD & \cite{zhu2020graph} & Combines hyperbolic distance $d_c$ with node feature $\mathbf{x}$ \\
        $\alpha_{ij} = \mathrm{sigmoid}(\mathrm{LeakReLU}(\mathbf{W}^T [\mathbf{x}_i; \mathbf{x}_j]) \times \frac{1}{\sqrt{d_i d_j}})$ & FI,SI& \cite{hgcn2019} &  Integrates node feature $\mathbf{x}$ with node degree information $d_i, d_j$ \\
        ${\alpha}_{ij}=\frac{w_\kappa \tilde{\kappa}_{ij}+w_f \tilde{f}_{ij}}{w_\kappa+w_f}$
        & FI,SI& \cite{yang2023kappahgcn} & Unifies node curvature information $w_\kappa$ with node feature$w_f$  \\
        \bottomrule
    \end{tabular}
    }
    \label{tab:aggration_summary}
    \vspace{-5pt}
\end{table*}

\subsubsection{Neighborhood Aggregation Mechanism}
\label{sec:weight_computation}
The determination of neighborhood weights is crucial in aggregating information effectively. Various strategies can be employed to calculate these weights, including utilizing structure information, feature information, hyperbolic distance, or a combination of these methods, which are summarized in Table~\ref{tab:aggration_summary}. Below, we delve into each of these strategies. Consider a central node \(i\), and we assume that \({N}(i)\) represents the immediate one-hop neighbor of node $i$, with \(j\) being an element of \(N(i)\)\footnote{Following the literature, we consider $i\in {N}(i)$.}.

(1) \textbf{Structure information}: The weight $\alpha_{ij}$ can be defined based on node degree~\cite{liu2019HGNN}:
\begin{equation}
    \alpha_{ij} = 1/\sqrt{\tilde{d_i}\tilde{d_j}},
    \label{equ:weight_structure}
\end{equation}
where $\tilde{d}_i = d_i+1$ and $d_i$ is the degree of node $i$. 
This method, derived from the Euclidean graph convolutional network~\cite{gcn2017}, provides a simplified approach to graph spectral convolution, primarily focusing on the topological information of the graph.

Additionally, the weight can also be defined using Ollivier Ricci curvature~\cite{ollivier2009ricci}, which describes the deviation of a local pair of neighborhoods from a "flat" case,
\begin{equation}
    \alpha_{ij} = \text{softmax}_{j \in N(i)} \left(\text{MLP}(\kappa_{i,j}) \right),
    \label{equ:weight_curvature}
\end{equation}
where MLP represents Multi-layer Perception and curvature $\kappa$ reflects the local structure of the node $i$, which is computed by optimal transport~\cite{yang2023kappahgcn}.

(2) \textbf{Feature information}~\cite{hgcn2019}: The weight $\alpha_{ij}$ using node feature is defined as follows:
\begin{equation}
{\alpha}_{ij}=\mathrm{softmax}_{j \in N(i)}(\mathrm{LeakReLU}(\mathbf{W}^T[\mathbf{x}_i^{\mathcal{T}_\mathbf{o}}||\mathbf{x}_j^{\mathcal{T}_\mathbf{o}}])).
\label{equ:weight_feature}
\end{equation}
This method computes the weight based on the hyperbolic features of nodes \(i\) and \(j\), drawing inspiration from the graph attention network~\cite{velivckovic2017graph}. 

(3) \textbf{Hyperbolic distance}~\cite{zhang2021hyperbolic}: The weight $\alpha_{ij}$ is formulated as:
\begin{equation}
\label{equ:weight_distance_hgat}
    \alpha_{ij} = \mathrm{softmax}_{j \in N(i)}\left(-d_\mathcal{H}^c(\mathbf{x}_i^{\mathcal{H}},\mathbf{x}_j^\mathcal{H})\right).
\end{equation}
This equation computes the weight based on the hyperbolic distance between nodes \(i\) and \(j\). The negative sign ensures that closer nodes have higher weights. Besides, Zhang et al.~\cite{zhang2021hyperbolic} also introduced $K$-head attention, which is defined as:
$$
\alpha_{ij} = \mathrm{softmax}_{j \in N(i)}\left(-\sqrt{\sum_{k=1}^K d_\mathcal{H}^c\left(\mathbf{x}^{\mathcal{H},k}_i, \mathbf{x}^{\mathcal{H},k}_j\right)}\right).
$$
Alternative approaches using hyperbolic distance include using squared distance~\cite{lgcn}:
\begin{equation}
\label{equ:weight_distance_lgcn}
    \alpha_{ij} = \mathrm{softmax}_{j \in N(i)}\left(-d_\mathcal{H}^c(\mathbf{x}_i^\mathcal{H},\mathbf{x}_j^\mathcal{H})^2\right).
\end{equation}
Similar to the previous method, the distance is squared, emphasizing the difference between close and distant nodes.
Another more complex form~\cite{HAT} is given as:
\begin{equation}
\alpha_{ij}=f(-\beta d_\mathcal{H}^c\left(\mathbf{x}_i^\mathcal{H}, \mathbf{x}_j^\mathcal{H})-\gamma\right),
\label{equ:weight_distance_hat}
\end{equation}
where $\beta$ and $\gamma$ are parameters that can be set manually or learned along with the training process, and $f$ can be $\mathrm{softmax(\cdot)}$. The hyperbolic distance is scaled by a factor \(\beta\) and shifted by \(\gamma\). Both \(\beta\) and \(\gamma\) can be learned or set manually. The function \(f\) can be a softmax function, ensuring the weights are normalized.

(4) \textbf{Hyperbolic cone}~\cite{tseng2023coneheads}: Hyperbolic cone emphasizing the difference between two points by the depth of their lowest common ancestor in a hierarchy defined by hyperbolic cones, the $\alpha_{ij}$ is computed by
\begin{equation}
{\alpha}_{ij}=f(d_\mathcal{H}^c(su{p}_{2}(\mathbf{x}_i^\mathcal{H},\mathbf{y}_i^\mathcal{H}),\mathcal{S}))
\end{equation}
where $\sup_2$ is root of the minimum height (literal lowest) cone that contains both $\mathbf{x}_i^\mathcal{H}$ and $\mathbf{y}_i^\mathcal{H}$, $\mathcal{S}$ is the root of all hierarchies, and $f$ is a user-defined monotonically increasing function.

(5) \textbf{Integration of feature information and hyperbolic distance}~\cite{zhu2020graph}: The $\alpha_{ij}$ is computed by
\begin{equation}
{\alpha}_{ij}=\mathrm{softmax}\left(\mathrm{LeakReLU}(\mathbf{W}^T[\mathbf{x}_i^{\mathcal{T}_\mathbf{o}}||\mathbf{x}_j^{\mathcal{T}_\mathbf{o}}])\times d_\mathcal{H}^c(\mathbf{x}_i^\mathcal{H},\mathbf{x}_j^{\mathcal{H}})\right).
\end{equation}
This approach combines feature attention and node distance. Compared with the pure feature method, it is modulated by the hyperbolic distance between the nodes, ensuring that both the node features and their relative positions in the hyperbolic space contribute to the weight.

(6) \textbf{Integration of feature and structure information}: The weight $\alpha_{ij}$ is computed by\footnote{From https://github.com/HazyResearch/hgcn}
\begin{equation}
{\alpha}_{ij}=\mathrm{sigmoid}\left(\mathrm{LeakReLU}(\mathbf{W}^T[\mathbf{x}_i^{\mathcal{T}_\mathbf{o}}||\mathbf{x}_j^{\mathcal{T}_\mathbf{o}}])\right)\times 1/\sqrt{\tilde{d}_i\tilde{d}_j}.
\end{equation}
This method integrates both structural information and feature attention, ensuring that both the node features and their connectivity in the graph contribute to the weight. In \cite{yang2023kappahgcn}, Yang et al.~\cite{yang2023kappahgcn} introduced another integration, which is formulated by:
$$
{\alpha}_{ij}=\frac{w_\kappa \tilde{\kappa}_{ij}+w_f \tilde{f}_{ij}}{w_\kappa+w_f},
$$
where $\tilde{\kappa}_{ij}$ and $\tilde{f}_{ij}$ is derived from Equation (\ref{equ:weight_curvature}) and (\ref{equ:weight_feature}), respectively.

\subsubsection{Overall View}
Having examined the individual components of hyperbolic GNNs, we now present a unified perspective on hyperbolic graph convolutional operations. In general, a unified hyperbolic graph convolutional layer can be formulated as:
\begin{equation}
\begin{array}{l}
\mathbf{h}_{i}^{\mathbf{\ell}, \mathcal{H}}=\left(\mathbf{W}^{\mathbf{\ell}} \otimes^{c_{\mathbf{\ell}-1},\mathcal{H}} \mathbf{x}_{i}^{\mathbf{\ell}-1, \mathcal{H}}\right) \oplus^{c_{\mathbf{\ell}-1},\mathcal{H}} \mathbf{b}^{\mathbf{\ell}}, \\
\mathbf{y}_{i}^{\mathbf{\ell}, \mathcal{H}}=\mathrm{AGG}^{c_{\mathbf{\ell}-1}}\left(\mathbf{h}^{\mathbf{\ell}, \mathcal{H}}\right)_{i}, \\
\mathbf{x}_{i}^{\mathbf{\ell}, \mathcal{H}}=\sigma^{{c_{\mathbf{\ell}-1},c_{\mathbf{\ell}}}}\left(\mathbf{y}_{i}^{\mathbf{\ell}, \mathcal{H}}\right).
\end{array}
\end{equation}
where $\otimes^{c,\mathcal{H}}$ and $\oplus^{c,\mathcal{H}}$ denote hyperbolic matrix multiplication and addition operations, $\mathrm{AGG}^c$ represents hyperbolic neighborhood aggregation, and $\sigma^{c,c'}$ denotes hyperbolic activation functions with potential curvature adjustment.

This unified formulation reveals a fundamental computational trade-off in hyperbolic GNN design. To reduce the computational overhead of repeated mappings between tangent space and hyperbolic manifold, some research works (e.g., Liu et al.~\cite{liu2019HGNN}) implement all three operations entirely within the tangent space. While this simplification reduces computational burden, it comes at the cost of decreased performance, as demonstrated by experimental results~\cite{zhu2020graph}, highlighting the tension between computational efficiency and geometric fidelity in hyperbolic neural architectures.

\textbf{Discussion and Summary.} Tangent space-based methods offer a practical compromise between Euclidean simplicity and hyperbolic expressiveness. These methods allow for relatively straightforward adaptation of existing deep learning techniques and operations (e.g., matrix multiplications, additions, non-linearities) that are well-established in Euclidean geometry, while inheriting the computational efficiency and mature optimization tools developed for Euclidean neural networks.

However, tangent space-based approaches face inherent limitations that stem from their hybrid nature. \textit{First}, the repeated use of exponential and logarithmic maps between the manifold and its tangent space incurs significant computational overhead, especially in deep architectures or large-scale graphs where these transformations accumulate across layers. \textit{Second}, since the tangent space provides only a local linear approximation of the curved manifold, repeated projections can introduce geometric distortion, particularly in high-curvature regions or when information is propagated across long geodesic distances.

In contrast, fully hyperbolic models preserve geometric fidelity throughout the computation by operating entirely within the curved space. However, this geometric consistency comes at the cost of mathematical and computational complexity, as these methods require redefining core operations (e.g., aggregation, attention mechanisms) in curved space using non-Euclidean Riemannian geometry. While more theoretically expressive, fully hyperbolic approaches tend to be mathematically intricate and computationally demanding, making stability and efficiency ongoing challenges in this research direction.

\subsection{Emerging Hyperbolic Graph Learning Paradigms}
\label{sec: Other Hyperbolic Graph Learning Paradigms}
Beyond the direct node embedding approaches discussed in Section~(\ref{sec: Hyperbolic Embedding}) and the message-passing GNN architectures explored in Section~(\ref{sec: GNN-based Hyperbolic Models}), the field of HGL has expanded to incorporate other powerful machine learning paradigms. These methods often leverage the unique geometric properties of hyperbolic space to enhance specific learning objectives, such as capturing global graph structures through diffusion processes or learning robust representations via self-supervised contrastive learning. 

\subsubsection{Diffusion-based Hyperbolic Representations}
\label{sec:Diffusion-based Hyperbolic Representations}
Diffusion models have achieved remarkable success in generating complex data by defining a forward noise process and learning to reverse it~\cite{song2019generative,ho2020denoising,song2020score}. Extending these models to hyperbolic space presents unique opportunities for graph data with inherent hierarchical structures.

HGDM~\cite{wen2024hyperbolic} addresses this challenge by adapting the diffusion framework to hyperbolic geometry using wrapped normal distributions—a natural generalization of Gaussian distributions to curved manifolds. The key insight is replacing Euclidean noise injection with hyperbolic perturbations that respect the manifold's geometry. The wrapped normal distribution on the Poincar\'e ball $\mathcal{B}_c^n$ is defined as:
\begin{equation}
\mathcal{N}_{\mathcal{B}_{c}^{n}}^{\mathrm{W}}(\mathbf{z}\mid\boldsymbol{\mu},\mathbf{\Sigma})=\mathcal{N}\left(\lambda_{\boldsymbol{\mu}}^{c}\log^c_{\boldsymbol{\mu}}(\mathbf{z})\mid\mathbf{0},\mathbf{\Sigma}\right)\left(\frac{\sqrt{c}d_{p}^{c}(\boldsymbol{\mu},\mathbf{z})}{\sinh\left(\sqrt{c}d_{p}^{c}(\boldsymbol{\mu},\mathbf{z})\right)}\right)^{n-1}
\end{equation}
where $\log^c_{\boldsymbol{\mu}}$ and $d_p^c$ denote the logarithmic map and Poincar\'e distance, respectively.

The framework employs two perturbation strategies adapted from Euclidean diffusion: Variance Exploding (VE) and Variance Preserving (VP), each corresponding to different stochastic differential equations in hyperbolic space. Both methods enable the generation of perturbed graphs $\bar{\mathbf{G}}_t$ from clean initial graphs $\bar{\mathbf{G}}_0$ at any time step $t$.

The training objective learns score functions $\mathbf{s}_{\theta,t}$ and $\mathbf{s}_{\phi,t}$ for node features and adjacency matrices respectively, minimizing the difference between predicted and true score functions. Graph generation is achieved through hyperbolic Predictor-Corrector (PC) sampling, which iteratively denoises from pure noise to structured graphs while maintaining hyperbolic geometric constraints. 

Building on the foundation of hyperbolic diffusion, HypDiff~\cite{fu2024hyperbolic} introduces a more computationally efficient approach by operating in latent space rather than directly on high-dimensional graph representations. This method employs a two-stage strategy: first training a hyperbolic autoencoder (comprising a hyperbolic encoder and Fermi-Dirac decoder), then learning the diffusion process within the resulting latent hyperbolic space.

The key innovation lies in designing an \textit{anisotropic} diffusion process that exploits hyperbolic geometry's inherent directional properties. Rather than treating all directions equally, the method constrains diffusion along both radial and angular dimensions to match the graph's structural patterns. The process begins by clustering nodes using hyperbolic k-means to identify natural groupings, then maps each node $\mathbf{h}_i$ in cluster $k$ to the tangent space at cluster center $\boldsymbol{\mu}_k$:
\begin{equation}
\mathbf{x}_{0_i}=\log_{\boldsymbol{\mu}_k}^c\left(\mathbf{h}_i\right)
\end{equation}
where $\mathbf{x}_{0_i}$ represents the initial tangent space coordinates of node $i$ obtained by mapping the hyperbolic node embedding $\mathbf{h}_i$ to the tangent space centered at cluster centroid $\boldsymbol{\mu}_k$.

The anisotropic diffusion then incorporates two geometric constraints. \textit{Angular constraints} ensure that generated nodes respect the directional structure implied by their cluster assignments, while \textit{radial constraints} preserve hyperbolic space's exponential volume growth. The unified diffusion process becomes:
\begin{equation}
\label{equ:fu latent}
\mathbf{x}_t=\sqrt{\overline{\alpha_t}}\mathbf{x}_0+\sqrt{1-\overline{\alpha_t}}\mathbf{z}+\delta\tanh\left[\sqrt{c}\lambda_o^ct/T_0\right]\mathbf{x}_0
\end{equation}
where $\mathbf{z}$ represents angle-constrained noise and the $\tanh$ term captures radial expansion effects with strength controlled by parameter $\delta$.

{\textbf{Discussion and Summary.}} These advances demonstrate that hyperbolic diffusion models can effectively generate graph-structured data while respecting geometric constraints. The progression from direct hyperbolic diffusion (HGDM) to geometrically-aware latent diffusion (HypDiff) illustrates how incorporating domain-specific geometric priors enhances both computational efficiency and generation quality. Latent space approaches not only reduce computational overhead but also enable more modular architectures for conditional generation tasks~\cite{yuan2025hyperbolic}.

\subsubsection{Hyperbolic Graph Contrastive Learning}
\label{sec: Hyperbolic Graph Contrastive Learning}
Contrastive learning has emerged as a powerful self-supervised paradigm that learns representations by maximizing agreement between different augmented views of the same instance while distinguishing them from negative examples. Traditional Euclidean graph contrastive learning relies on structural perturbations (e.g., node dropping, edge perturbation, subgraph sampling) or feature-level augmentations to create diverse views~\cite{you2020graph,zhu2021graph,yu2022graph}. However, extending contrastive learning to hyperbolic space introduces unique opportunities and challenges, as the curved geometry fundamentally alters both the augmentation strategies and the underlying contrastive objectives.

The key insight is that hyperbolic geometry itself can serve as a source of meaningful augmentations, while the exponential growth properties of hyperbolic space require careful consideration of uniformity and alignment objectives to prevent representation collapse—a critical challenge that is exacerbated in curved manifolds.

{\textbf{Geometric Augmentation Strategies.}} A distinctive approach leverages the mathematical isomorphism between different hyperbolic models as a natural augmentation strategy. HGCL~\cite{liu2022enhancing} exemplifies this by treating the Lorentz model and Poincar\'e ball as complementary views of the same underlying graph structure. Rather than artificially perturbing the graph, this method exploits the fact that these models offer different computational and geometric perspectives on identical hyperbolic relationships. The contrastive objective aligns representations across these dual geometric views, effectively combining the computational advantages of different hyperbolic models with self-supervised learning. This geometric augmentation strategy extends beyond traditional perturbation-based methods by utilizing the inherent mathematical structure of hyperbolic space itself.

{\textbf{Representation Quality Preservation.}} A fundamental challenge in contrastive learning is dimensional collapse, where representations converge to a lower-dimensional subspace, severely limiting expressive capacity. This problem also exists in hyperbolic space. Zhang et al.~\cite{zhang2023alignment} propose HyperGCL to decompose the contrastive objective into alignment and uniformity components. HyperGCL introduces leaf-level uniformity, encouraging diversity among leaf nodes in hierarchical structures, and height-level uniformity, maintaining diversity across different hierarchy levels. By carefully balancing these mechanisms, the approach preserves both the hierarchical structure inherent in hyperbolic embeddings and the representational diversity required for effective downstream performance.

\textbf{Hierarchical Structure Exploitation.} HCGR~\cite{guo2021hcgr} applies hyperbolic contrastive learning to session-based recommendation by converting user sessions into directed graphs in Lorentz hyperbolic space. The method uses adaptive hyperbolic attention to capture sequential dependencies and employs hyperbolic geodesic distances in the contrastive objective to preserve hierarchical relationships. The exponential distance scaling naturally accommodates the hierarchical structure of user behavior patterns and item taxonomies.

{\textbf{Discussion and Summary.}} Hyperbolic contrastive learning represents a significant departure from traditional approaches by incorporating geometric structure directly into the learning objective. The ability to simultaneously preserve local neighborhoods and global hierarchical relationships enables more nuanced representations that capture both fine-grained similarities and coarse-grained structural patterns. These methods demonstrate superior performance on tasks requiring hierarchical understanding, such as link prediction in knowledge graphs and node classification in taxonomic networks.

\subsection{Optimization}
The optimization of hyperbolic representation learning presents unique challenges due to the curved geometry of the underlying space. The choice of optimization strategy fundamentally affects both computational efficiency and model expressiveness, leading to three distinct approaches that differ in where parameters reside and how gradients are computed.

\textbf{Riemannian Optimization.} When parameters such as node embeddings are directly optimized within hyperbolic space, standard Euclidean optimizers become inapplicable since they assume flat geometry and may violate manifold constraints. Riemannian optimization methods address this by adapting gradient-based algorithms to curved manifolds. Key approaches include Riemannian SGD (RSGD)~\cite{bonnabel2013stochastic}, which projects gradients onto the tangent space and uses exponential maps to update parameters while maintaining manifold constraints, and Riemannian variants of adaptive methods such as Riemannian Adam~\cite{becigneul2018riemannian} and Riemannian Adagrad~\cite{becigneul2018riemannian}. These methods ensure that parameter updates remain on the manifold by computing Riemannian gradients and using retraction operations, effectively preserving the geometric structure during optimization~\cite{nickel2017poincare,nickel2018learning,sun2021hgcf,yang2022hrcf,yang2022hicf}.

\textbf{Euclidean Optimization.} An alternative approach maintains all learnable parameters in Euclidean space while incorporating hyperbolic geometry through forward-pass computations. In this strategy, parameters are optimized using conventional optimizers such as Adam~\cite{kingma2014adam} or Adagrad~\cite{duchi2011adaptive}, while hyperbolic operations (exponential/logarithmic maps, geodesic distances, hyperbolic aggregations) are applied as geometric transformations during inference~\cite{hgcn2019,yang2023kappahgcn,lgcn}. This approach provides significant practical advantages: compatibility with existing deep learning frameworks, numerical stability due to well-conditioned Euclidean gradients, and computational efficiency by avoiding complex Riemannian computations. However, this strategy may constrain the optimization dynamics to Euclidean geometry, potentially limiting the model's ability to fully exploit hyperbolic space's exponential capacity for representing hierarchical structures.

\textbf{Hybrid Optimization.} A third strategy combines both approaches by selectively applying Riemannian optimization to critical hyperbolic components while using Euclidean optimization for auxiliary parameters. For example, node embeddings may be optimized in hyperbolic space using Riemannian methods, while attention weights, transformation matrices, or bias terms remain in Euclidean space and are updated with standard optimizers~\cite{chen2021fully,yang2024hypformer}. This hybrid approach seeks to balance the geometric fidelity of Riemannian optimization with the computational efficiency and stability of Euclidean methods, allowing practitioners to optimize the trade-off between model expressiveness and computational cost based on specific application requirements.

\textbf{Discussion and Summary.} The choice of optimization strategy in HGL involves a fundamental trade-off between geometric fidelity and computational practicality. Riemannian optimization preserves the intrinsic geometry of hyperbolic space but requires specialized algorithms and careful numerical implementation. Euclidean optimization offers computational simplicity and framework compatibility but may limit the model's ability to fully exploit hyperbolic geometry's representational advantages. Hybrid approaches provide a middle ground, enabling selective application of geometric constraints where they matter most.

\section{Applications}
\label{sec4}
HGLs have demonstrated remarkable potential across a wide range of application domains. The inherent ability of hyperbolic spaces to model complex, hierarchical, and scale-free structures with minimal distortion makes them uniquely suited for real-world data, which often exhibits such non-Euclidean characteristics. The negative curvature of hyperbolic spaces naturally accommodates exponential expansion, making them particularly well-suited for modeling complex graphs that are difficult to capture in Euclidean geometry, such as those found in recommender systems, KGs, and biological networks. This section provides a comprehensive overview of the key application areas where HGLs have demonstrated superior performance and yielded novel insights compared to their Euclidean counterparts. Table~\ref{tab:summary of applications} provides an overview of some advanced applications discussed in this section.

\begin{table*}[t]
\centering
\caption{Summary of the advanced applications in the hyperbolic space. Here, ’Mixed’ indicates methods that combine multiple geometries (e.g., Euclidean, elliptic, and hyperbolic) within the same model.}
\begin{tabular}{cccccc}
\hline
\multicolumn{1}{l}{}  & Category                                     & Year                 & Paper                & Geometry               & Method                                          \\ \hline\hline
\multicolumn{1}{c|}{\multirow{22}{*}{\rotatebox{90}{Recommendation System}}} & \multirow{9}{*}{4.1.1 Collaborative Filtering}    & 2020   & HyperML~\cite{DBLP:conf/wsdm/TranT0CL20}     & Poincar\'e    & Distance-based Embedding     \\ 
\multicolumn{1}{c|}{}       &           & 2021                 & HGCF~\cite{sun2021hgcf}                 & Lorentz                      & Tangent-based GNN                               \\ 
\multicolumn{1}{c|}{}       &           & 2021                 & LGCF~\cite{wang2021fully}                 & Poincar\'e,Klein                    & Fully hyperbolic GNN                            \\ 
\multicolumn{1}{c|}{}       &           & 2022                 & HRCF~\cite{yang2022hrcf}                 & Lorentz                      & Tangent-based GNN                               \\ 
\multicolumn{1}{c|}{}       &           & 2022                 & HICF~\cite{yang2022hicf}                 & Lorentz                      & Tangent-based GNN                               \\ 
\multicolumn{1}{c|}{}       &           & 2022                 & HNCR~\cite{li2022hyperbolic}                 & Lorentz                      & Tangent-based GNN                               \\ 
\multicolumn{1}{c|}{}       &           & 2024                 & HVACF~\cite{shimizu2024fashion}                 & Poincar\'e                      & Tangent-based GNN, CV                 \\
\multicolumn{1}{c|}{}       &       & 2025                 & HDRM~\cite{yuan2025hyperbolic}                 & Poincar\'e                      & Tangent-based GNN, Graph Latent Diffusion    \\  
\multicolumn{1}{c|}{}       &       & 2025                 & HyperLLM~\cite{cheng2025large}                 & Poincar\'e,Lorentz                      & Fully hyperbolic GNN, LLM    \\ \cline{2-6} \cline{2-6}
\multicolumn{1}{c|}{}        & \multirow{5}{*}{4.1.2 KG-enhanced Recommendation}     & 2021      & KBHP~\cite{tai2021knowledge}                 & Lorentz,Klein          & Tangent-based GNN                \\ 
\multicolumn{1}{c|}{}   &      & 2022                 & LKGR~\cite{chen2022modeling}                 & Lorentz                      & Tangent-based GNN                               \\ 
\multicolumn{1}{c|}{}   &      & 2022                 & HAKG~\cite{du2022hakg}                 & Poincar\'e                      & Tangent-based GNN                               \\ 
\multicolumn{1}{c|}{}   &     & 2023                 & LECF~\cite{huang2024lorentz}                 & Lorentz                      & Fully hyperbolic GNN                            \\ 
\multicolumn{1}{c|}{}   &     & 2024                 & HGCH~\cite{zhang2024hgch}                 & Poincar\'e                      & Tangent-based GNN                            \\ \cline{2-6}
\multicolumn{1}{c|}{}& \multirow{3}{*}{4.1.3 Social-aware Recommendation}       & 2021         & HyperSoRec~\cite{wang2021hypersorec}           & Lorentz             & Tangent-based GNN              \\ 
\multicolumn{1}{c|}{}   &     &   2022              &   HSR~\cite{li2022hsr}         &      Poincar\'e                 & Tangent-based GNN                      \\ 
\multicolumn{1}{c|}{}   &     &   2023              &   HGSR~\cite{yang2023hyperbolic}         &      Lorentz        & Tangent-based GNN                      \\ \cline{2-6}
\multicolumn{1}{c|}{}& \multirow{5}{*}{4.1.4 Session-based Recommendation}     & 2021    & HCGR~\cite{guo2021hcgr}        & Lorentz                      & Tangent-based GNN, Contrastive Learning         \\ 
\multicolumn{1}{c|}{}    &        & 2022                 & H$^{2}$SeqRec~\cite{li2021hyperbolic}             & Lorentz                      & Tangent-based GNN, Hypergraph                   \\ 
\multicolumn{1}{c|}{}    &        & 2023                 & HADCG~\cite{su2023enhancing}                & Lorentz                      & Tangent-based GNN, Contrastive Learning         \\ 
\multicolumn{1}{c|}{}    &        & 2024                 & MIHRN~\cite{liu2024enhancing}             & Lorentz                      & Tangent-based GNN, Hypergraph         \\ 
\multicolumn{1}{c|}{}    &        & 2025                 & HMamba~\cite{zhang2025hmamba}             & Lorentz                      & Tangent-based GNN         \\ \hline\hline
\multicolumn{1}{c|}{\multirow{12}{*}{\rotatebox{90}{Knowledge Graph}}} & \multirow{7}{*}{4.2.1 Transformation-based Method} & 2019   & MuRP~\cite{DBLP:conf/nips/BalazevicAH19}  & Poincar\'e   & Distance-based Embedding  \\ 
\multicolumn{1}{c|}{}&  & 2019                 & AttH~\cite{ChamiKG}                 & Poincar\'e                      & Angle-based Embedding (rotation and reflection) \\ 
\multicolumn{1}{c|}{}&  & 2019                 & HyperKG~\cite{kolyvakis2019hyperkg}              & Poincar\'e                      & Distance-based Embedding                        \\ 
\multicolumn{1}{c|}{}&  & 2021                 & HBE~\cite{pan2021hyperbolic}                  & Extended Poincar\'e & Angle-based Embedding (polar coordinate)        \\ 
\multicolumn{1}{c|}{}&  & 2021                 & ConE~\cite{bai2021modeling}                 & Poincar\'e                      & Angle-based Embedding (Cone)                    \\ 
\multicolumn{1}{c|}{}&  & 2022                 & GIE~\cite{cao2022geometry}                  & Mixed                  & Distance-based Embedding                        \\ 
\multicolumn{1}{c|}{}&  & 2024                 & LorentzKG~\cite{fan2024enhancing}                  & Lorentz       & Distance-based Embedding (Fully hyperbolic)        \\ \cline{2-6}
\multicolumn{1}{c|}{}& \multirow{5}{*}{4.2.2 GNN-based Method}    & 2020     & HyperKA~\cite{sun2020knowledge}     & Poincar\'e         & Tangent-based GNN                               \\ 
\multicolumn{1}{c|}{}&      & 2021                 & H$^{2}$E~\cite{wang2021knowledge}                  & Poincar\'e                      & Tangent-based GNN                               \\ 
\multicolumn{1}{c|}{}&      & 2021                 & M$^{2}$GNN~\cite{wang2021mixed}                & Mixed                  & Tangent-based GNN                               \\ 
\multicolumn{1}{c|}{}&      & 2023                 & FFHR~\cite{shi2023ffhr}                 & Poincar\'e                      & Fully hyperbolic GNN                            \\ 
\multicolumn{1}{c|}{}&      & 2024                 & MGTCA~\cite{shang2024mixed}                & Mixed                  & Tangent-based GNN                               \\ \hline\hline
\multicolumn{1}{c|}{\multirow{8}{*}{\rotatebox{90}{Biograph}}} & \multirow{5}{*}{4.3.1 Bioinformatics Representation}   & 2021     & HiG2Vec~\cite{kim2021hig2vec}  & Poincar\'e    & Distance-based Embedding       \\ 
\multicolumn{1}{c|}{} &   & 2021                 & HRGCN+~\cite{wu2021hyperbolic}               & Poincar\'e                      & Tangent-based GNN                               \\ 
\multicolumn{1}{c|}{} &   & 2020                 & SHDE~\cite{yu2020semi}                 & Lorentz                      & Tangent-based GNN, VAE                          \\ 
\multicolumn{1}{c|}{} &   & 2021                 & scPhere~\cite{ding2021deep}              & Poincar\'e                      & Tangent-based GNN, VAE                          \\ 
\multicolumn{1}{c|}{} &   & 2024                 & CPM~\cite{bhasker2024contrastive}                  & Poincar\'e                      & Tangent-based GNN, Contrastive Learning         \\ \cline{2-6}
\multicolumn{1}{c|}{}& \multirow{3}{*}{4.3.2 Molecular Generation}    & 2024    & HGDM~\cite{wen2024hyperbolic}                 & Poincar\'e                      & Graph Diffusion                                 \\ 
\multicolumn{1}{c|}{}    &              & 2024                 & HypDiff~\cite{fu2024hyperbolic}              & Lorentz                      & Graph Latent Diffusion                          \\ 
\multicolumn{1}{c|}{}    &              & 2025                 & GGBall~\cite{bu2025ggball}              & Poincar\'e                      & Graph Latent Diffusion                          \\ 
\hline
\end{tabular}
\label{tab:summary of applications}
\end{table*}

\subsection{Recommender Systems}
Recommender systems model user preferences through bipartite graphs where users and items form nodes connected by interaction edges. The power-law distributions prevalent in user-item networks align naturally with hyperbolic geometry's exponential capacity, making it well-suited for capturing hierarchical preference structures and complex interaction patterns. This section examines hyperbolic applications across four recommendation paradigms: collaborative filtering, knowledge graph-enhanced recommendation, social-aware recommendation, and session-based recommendation. 

\subsubsection{Collaborative Filtering}
Hyperbolic collaborative filtering leverages the hierarchical nature of user-item networks to model preference patterns more effectively than Euclidean approaches. Research has evolved along three main directions.

\textbf{Framework Optimization.} HyperML~\cite{DBLP:conf/wsdm/TranT0CL20} pioneered hyperbolic recommender systems by incorporating hyperbolic distances into margin ranking losses. Subsequent works enhanced architectures: HGCF~\cite{sun2021hgcf} employs graph convolutions for long-range relationship modeling, while LGCF~\cite{wang2021fully} introduces fully hyperbolic frameworks using Einstein midpoint aggregation in Klein space. HDRM~\cite{yuan2025hyperbolic} integrates diffusion models with anisotropic processes to preserve topological structures.

\textbf{Loss Function Enhancement.} Specialized loss functions better exploit hyperbolic geometry's properties. HRCF~\cite{yang2022hrcf} implements hyperbolic regularization with root alignment and origin-aware penalties to prevent embedding collapse. HICF~\cite{yang2022hicf} enhances margin ranking through hyperbolic-aware learning and strategic negative sampling, improving recommendations for both popular and niche items.

\textbf{Semantic Enhancement.} Beyond interaction modeling, semantic approaches capture implicit user-item relationships. HNCR~\cite{li2022hyperbolic} constructs semantic graphs from co-occurrence frequencies using nearest neighbor approaches. Recent methods like HVACF~\cite{shimizu2024fashion} and HyperLLM~\cite{cheng2025large} incorporate visual features and pre-trained models to derive richer item representations. 

\subsubsection{Knowledge Graph-Enhanced Recommendation}
Knowledge graph-enhanced systems leverage explicit semantic relationships to address cold-start problems and sparse interactions. Methods divide into two categories based on modeling strategies.

\textbf{Unified Methods.} These integrate users, items, and KG entities into unified graph structures. KBHP~\cite{tai2021knowledge} adapts RippleNetwork to hyperbolic space with specialized scoring functions and multi-hop entity aggregation. LKGR~\cite{chen2022modeling} employs attention mechanisms on Lorentz manifolds with distinct propagation strategies for heterogeneous information encoding.

\textbf{Separate Modeling Methods.} These treat KGs as auxiliary semantic sources rather than unified components. HAKG~\cite{du2022hakg} constructs dual item embeddings through separate aggregation strategies, using gating mechanisms for adaptive integration and cone constraints. LECF~\cite{huang2024lorentz} provides fully hyperbolic modeling with Lorentz-equivariant transformations for joint attribute and embedding refinement.

\subsubsection{Social-Aware Recommendation}
Social relationships significantly influence user preferences, motivating integration of social network information with collaborative filtering signals.

HSR~\cite{li2022hsr} pioneered social recommendation in hyperbolic space through weighted HGNNs modeling user-item and user-user relationships. HyperSoRec~\cite{wang2021hypersorec} captures multi-aspect interactions via learnable relations for specific user-item pairs. HGSR~\cite{yang2023hyperbolic} introduces social embedding pre-training to capture hierarchical social structures, combining user-item graph convolutions with social diffusion modeling.

\subsubsection{Session-Based Recommendation}
Session-based recommendation predicts user actions from anonymous, sequential interaction sessions, presenting unique temporal modeling challenges for hyperbolic approaches.

HCGR~\cite{guo2021hcgr} first applied hyperbolic modeling to session-based recommendation, using hyperbolic self-attention on sequence graphs with contrastive learning for hierarchical behavior extraction. H$^2$SeqRec~\cite{li2021hyperbolic} employs hyperbolic hypergraph convolutions for high-order relationship modeling. MIHRN~\cite{liu2024enhancing} integrates multi-interest mechanisms with hypergraph networks for diverse user interest representation. HADCG~\cite{su2023enhancing} addresses collaborative signal limitations through dual clustering modules with intra-session and inter-session clustering strategies.

\subsection{Knowledge Graphs}
Knowledge graphs represent structured information as entity-relation-entity triples, encoding real-world facts in graph form. Large-scale KGs exhibit scale-free distributions and hierarchical organization patterns, making hyperbolic geometry well-suited for learning embeddings that preserve these structural properties while maintaining low dimensionality.

\subsubsection{Geometric Transformation Methods}
Geometric transformation approaches model KG relations through spatial operations in embedding space, extending classical translation and rotation paradigms to hyperbolic manifolds.

\textbf{Hyperbolic Extensions of Classical Models.} MuRP~\cite{DBLP:conf/nips/BalazevicAH19} adapts translational models to the Poincar\'e ball using relation-specific M\"obius transformations for modeling concurrent hierarchies. AttH~\cite{ChamiKG} introduces learnable relation-specific curvatures with hyperbolic reflections and rotations, combining them through attention mechanisms. HyperKG~\cite{kolyvakis2019hyperkg} enhances translational approaches for quasi-chain Datalog rule representation.

\textbf{Addressing Geometric Limitations.} Standard hyperbolic embeddings suffer from boundary crowding issues. HBE~\cite{pan2021hyperbolic} addresses this through polar coordinate representations where radius encodes hierarchy and angle differentiates same-level entities. ConE~\cite{bai2021modeling} employs hyperbolic cones for inclusion relations, modeling hierarchical relations through cone containment constraints and non-hierarchical relations via cone rotations.

\textbf{Multi-Space Approaches.} Recognizing that single geometric spaces cannot capture all KG topologies, GIE~\cite{cao2022geometry} integrates Euclidean, hyperbolic, and spherical spaces into unified frameworks. LorentzKG~\cite{fan2024enhancing} avoids projection distortions through fully hyperbolic Lorentz model embeddings.

\subsubsection{Graph Neural Network Methods}
GNN-based approaches leverage neighborhood information for enhanced generalization and long-range dependency modeling in hyperbolic spaces.

\textbf{Basic Hyperbolic GNN Approaches.} HyperKA~\cite{sun2020knowledge} employs hyperbolic GNNs for multi-granular knowledge association capture. H$^2$E~\cite{wang2021knowledge} introduces attentive context aggregation for hierarchical relation preservation in KG completion tasks.

\textbf{Geometric Fidelity Improvements.} Most methods rely on tangent space approximations that introduce distortions. FFHR~\cite{shi2023ffhr} performs message passing directly in the Poincar\'e model, introducing generalized hyperbolic inner products to replace distance-based scoring for better relational pattern modeling.

\textbf{Hybrid Geometric Approaches.} Recent work explores mixed-geometry solutions for diverse KG topologies. M$^{2}$GNN~\cite{wang2021mixed} constructs learnable curvature product manifolds for enhanced structural expressiveness. MGTCA~\cite{shang2024mixed} integrates geometry-aware attention mechanisms for improved multi-structural KG aggregation, demonstrating complementary approaches to handling topological diversity.

The evolution of hyperbolic knowledge graph methods demonstrates a clear progression from basic geometric adaptations to sophisticated multi-space frameworks. While geometric transformation methods provide intuitive extensions of classical KG embeddings, GNN-based approaches offer superior scalability and expressiveness through neighborhood aggregation.

\subsection{Bioinformatics and Molecular Graphs}
Molecules are naturally represented as graphs, with nodes representing atoms and edges representing chemical bonds. The prevalence of hierarchical structures in biological systems—from protein folding patterns to gene regulatory networks—makes hyperbolic geometry particularly well-suited for capturing these complex relationships. Recent advances in hyperbolic graph neural networks have demonstrated significant improvements in biological applications, leveraging the exponential growth properties of hyperbolic space to model hierarchical biological structures more effectively than traditional Euclidean approaches.

\subsubsection{Bioinformatics Graphs Representation}
\textbf{Gene Ontology and Functional Analysis.}
HiG2Vec~\cite{kim2021hig2vec} embeds Gene Ontology (GO) terms and genes into a shared hyperbolic space via a two-step process: it first maps the GO DAG into the Poincaré ball, then positions genes based on GO annotations to ensure semantic alignment with related GO terms. This approach leverages the natural tree-like structure of GO hierarchies, where the exponential volume growth of hyperbolic space naturally accommodates the branching complexity of biological ontologies.

\textbf{Drug Discovery and Molecular Property Prediction}. HRGCN+~\cite{wu2021hyperbolic} combines molecular graphs and molecular descriptors for drug discovery, allowing medicinal chemists to understand models at both the atom and descriptor levels while aiding in the extraction of hidden information. SHDE~\cite{yu2020semi} proposes learning molecular embedding through the hyperbolic VAE framework to discover new side effects and re-position drugs. These approaches demonstrate how hyperbolic geometry can capture the hierarchical nature of molecular structures, from atomic interactions to pharmacological properties.

\textbf{Single-cell Analysis and Genomics.} 
scPhere~\cite{ding2021deep} is a scalable generative model that embeds high-dimensional scRNA-seq data into a low-dimensional space by combining VAEs with hyperspherical and hyperbolic geometry. 
Existing hyperbolic methods prioritize local structure but sacrifice global accuracy and efficiency. To address this, CPM~\cite{bhasker2024contrastive} combines contrastive learning and hyperbolic space methods, demonstrating superior efficiency and scalability. These methods effectively model the hierarchical organization of cellular differentiation processes and gene regulatory networks.

\subsubsection{Molecular Generation}
The field of molecular generation has witnessed significant advances through hyperbolic generative models.

Molecular generation in hyperbolic space is typically achieved through graph-based models such as normalizing flows~\cite{bose2020latent} and GANs~\cite{qu2022hyperbolic}.
Recently, diffusion models~\cite{song2020score,song2019generative} have attracted considerable attention due to their strong generative capabilities and well-grounded mathematical foundations. 
HGDM~\cite{wen2024hyperbolic} is the first method to combine diffusion models with hyperbolic representations for graph and molecular generation. By leveraging the wrapped normal distribution in hyperbolic space, the model directly performs node-level diffusion to generate both graph structure and node features. 
To improve controllability and generation quality, HypDiff~\cite{fu2024hyperbolic} adopts the latent diffusion paradigm, replacing the original data space with a low-dimensional latent space to reduce complexity and enable more efficient molecular graph generation. 
Building on this foundation, GGBall~\cite{bu2025ggball} enhances the autoencoder module and introduces the Poincaré Diffusion Transformer.

\subsection{Other Application Scenarios}
This section highlights additional application scenarios of HGLs across diverse domains beyond those previously discussed. 
For skeleton-based action recognition, Peng et al.~\cite{peng2020mix} designed a hyperbolic spatial-temporal GCN that combines several dimensions on the manifold and provides an effective technique to explore the dimension for each ST-GCN layer. 
For quantitative trading and investment decision-making, Sawhney et al.~\cite{sawhney2021exploring} modeled the inter-stock relations by HGNN and developed a stock model, HyperStockGAT, which constructs the stock graph by the relation in Wikidata and their industry information. 

\section{Challenges and Opportunities}
\label{sec5}
Despite the rapid progress of HGL in recent years, several open challenges remain that require more effective solutions. The existing survey~\cite{peng2021hyperbolic}  discussed several open problems that are also shared by HGL. In this section, we further summarize several challenges in the HGL community, which also provide opportunities for future study.

\subsection{Challenges I: Complex Structures}
In graph representation learning, hyperbolic space is emerging as a potential alternative. The most noticeable advantages are credited to its exponential volume growth property, which makes this space much more embeddable than Euclidean space, especially for datasets with implicit tree-like layouts. However, real-world networks often exhibit mixed geometries, where certain regions are tree-like, while others are flat or cyclic, resulting in structural heterogeneity. Directly embedding a graph with intricate layouts into the hyperbolic manifold inevitably leads to structural inductive biases and distortions. 

\textit{Opportunities:}\hspace{1em} 
To date, there have been some preliminary attempts, but there is still considerable potential for improvement. A common strategy is to integrate Euclidean and/or spherical spaces to complement the expressive capacity of hyperbolic space~\cite{cao2022geometry,wang2021mixed,shang2024mixed}. For example, GIL~\cite{zhu2020graph} utilizes hyperbolic space and Euclidean space interactively and places different weights on two branches to cope with intricate complex graph structures. 
SelfMGNN~\cite{sun2021self} resorts to a mixed-curvature space via the Cartesian product. 
On the other hand, ACE-HGNN~\cite{fu2021ace} attempts to learn an optimal curvature to model the tree-like graph with different hyperbolicity via a multi-agent reinforcement learning framework.
HGCL~\cite{liu2022enhancing} enhances the modeling of HGNN by contrastive learning.
These methods generally need to create multiple branches, which unavoidably increases the computational complexity to a certain extent. 
Furthermore, most of the above methods adopt global modeling strategies. Incorporating local geometric indicators, such as Ricci curvature~\cite{ni2019community_ricci_flow,topping2021understanding}, may offer more efficient and adaptive solutions.

\subsection{Challenge II: Geometry-aware Learning}
Though HGL has made noticeable achievements, most of the efforts mainly focus on how to generalize graph representations into hyperbolic space by properly designing the transformation operations among the spaces. 
However, the optimization objectives—such as the use of cross-entropy loss in HGCN~\cite{hgcn2019} and LGCN~\cite{lgcn}—are often borrowed from Euclidean settings, and typically fail to account for the intrinsic properties of hyperbolic geometry.  
On the other hand, hyperbolic space is curved, that is, locations closer to the origin are flatter and relatively narrow, whereas regions further from the origin are broader, and this property is seldom considered when designing hyperbolic models.

\textit{Opportunities:} A key direction is to align the optimization objective with the underlying hyperbolic geometry, rather than relying on Euclidean-style loss functions. This constitutes the \textit{first} challenge worth addressing. 
Recently, HRCF~\cite{yang2022hrcf} presented a geometrically aware hyperbolic recommender system, paving the path for the community.  
The \textit{second} opportunity lies in making the learning process itself geometry-aware~\cite{iizuka2021dynamic}, keeping the awareness of node position rather than being equipped with a geometry-unconscious optimization target, like the constant margin in metric learning. 
\textit{Last}, due to the exponential expansion of hyperbolic space, regions far from the origin offer abundant capacity for distributing samples with minimal overlap~\cite{yang2023hyperbolic1}. Intuitively, by encouraging the model to preserve the inherent data structure and pushing the overall embedding far away from the origin, the model representation ability could then be largely improved. 

\subsection{Challenges III: Trustworthiness of HGL}
While HGL has proven effective for hierarchical graph representation, several trustworthiness issues remain unresolved. The rationale for the superiority of hyperbolic space is unclear, specifically whether performance gains arise from the improved representation of high-level nodes, tail nodes, or both. Additionally, the conditions under which hyperbolic models outperform Euclidean counterparts are poorly understood, with generalization error and robustness properties remaining largely underexplored.

\textit{Opportunities:} 
These challenges present significant research opportunities for advancing the field. Initial investigations like HSCML~\cite{zhang2021we} have empirically studied hyperbolic recommender systems, demonstrating the potential for systematic trustworthiness analysis in HGL tasks. Theoretical advances by Suzuki et al.~\cite{suzuki2021HOE} provide a promising foundation, showing that hyperbolic ordinal embedding (HOE) achieves generalization error bounds that are at most exponential with respect to embedding radius while representing trees more effectively than linear graph embedding~\cite{suzuki2021GraphEmbedding}.
Building on these foundations, researchers can develop comprehensive theoretical frameworks to characterize when and why hyperbolic models excel, potentially leading to principled model selection criteria and performance guarantees. The development of robustness metrics specifically tailored to hyperbolic geometry could enable more reliable HGL systems, while interpretability tools for hyperbolic embeddings would provide insights into the geometric properties that drive performance improvements.

\subsection{Challenges IV: Scalable HGL}
Despite significant performance improvements, HGL faces substantial scalability challenges due to higher computational costs compared to Euclidean counterparts. Tangent space-based models particularly suffer from expensive exponential and logarithmic map computations, which become prohibitive in high dimensions. With real-world graphs often containing millions of nodes and edges under limited computational resources, scaling HGL to large graphs remains a critical open challenge.

\textit{Opportunities:} 
These scalability challenges present exciting opportunities for methodological innovation, which can be achieved from two complementary aspects. 
\textit{\textbf{Scalable GNN Architectures.}} Various GNN scalability techniques offer promising adaptation pathways for hyperbolic frameworks. Decoupling feature transformation from propagation~\cite{sgc,rossi2020sign} could significantly reduce computational overhead while preserving hyperbolic geometric properties. Graph sampling techniques~\cite{zeng2019graphsaint,hamilton2017SAGE,huang2021scaling} enable efficient processing of large-scale hierarchical structures by intelligently selecting representative subgraphs that maintain hyperbolic relationships. These approaches can be directly integrated into existing hyperbolic GNN architectures without compromising their geometric advantages.
\textit{\textbf{Efficient Hyperbolic Computation.}} Recent breakthroughs in hyperbolic architectures address computational bottlenecks directly. The introduction of Hypformer~\cite{yang2024hypformer} represents a major advancement, achieving the first hyperbolic Transformer with linear self-attention complexity, enabling processing of billion-scale graph data for the first time. Meta-learning approaches such as H-GRAM~\cite{choudhary2023hyperbolic} address scalability through transferable hyperbolic representations, enabling faster learning on new large subgraphs with disjoint node sets. Additionally, approximation algorithms specifically designed for hyperbolic operations could dramatically reduce the computational burden of exponential and logarithmic maps without sacrificing representation quality.

\subsection{Challenges V: Integration With Foundation Models}
The emergence of foundation models, especially the LLMs, has revolutionized AI across multiple domains~\cite{grattafiori2024llama,yang2025qwen3,liu2025survey,radford2021learning,brandes2022proteinbert}. 
However, integrating these powerful foundation models with hyperbolic graph learning presents several challenges.
\textit{First}, geometric incompatibility poses a fundamental barrier, as foundation models are designed and pre-trained entirely in Euclidean space~\cite{yang2024hyperbolic}. This misalignment between pre-trained Euclidean embeddings and hyperbolic representation spaces causes significant performance degradation~\cite{mandica2025hyperbolic}. 
\textit{Second}, scalability becomes prohibitive when extending hyperbolic methods to billion-parameter models due to computational overhead from additional mapping operations and Riemannian computations~\cite{mandica2025hyperbolic}. 
\textit{Third}, training instability emerges during pre-training or fine-tuning in hyperbolic space~\cite{mishne2023numerical,mandica2025hyperbolic}, where naive application of exponential and logarithmic maps leads to cancellation effects that compromise hyperbolic modeling capabilities~\cite{yang2024hyperbolic}.

\textit{Opportunities:} Despite these challenges, the integration offers substantial advantages. Enhanced hierarchical representation learning naturally captures tree-like structures inherent in language~\cite{dhingra2018embedding,le2019inferring,chen2024hyperbolic,he2025helm}, while improved embedding efficiency enables large-scale data processing with reduced dimensionality while preserving complex relationships~\cite{sala2018representation,nickel2017poincare,yang2022hyperbolic_htgn}. The exponential growth properties of hyperbolic space also benefit few-shot learning scenarios~\cite{zhang2022hyperbolic,wang2024metahkg}.
Multimodal hyperbolic learning represents another promising direction, as demonstrated by recent hyperbolic vision-language models~\cite{mandica2025hyperbolic,desai2023hyperbolic,pal2024compositional}, which capture hierarchical relationships across modalities while providing uncertainty quantification through embedding norms or cones. Additionally, recommender systems could leverage LLM-enhanced hyperbolic embeddings that combine semantic understanding with geometric structure~\cite{ma2024harec}, though comprehensive empirical validation remains needed.
Future research could focus on developing efficient~\cite{yang2024hypformer}, fully~\cite{he2025helm}, or hybrid architectures that bridge Euclidean and hyperbolic spaces, creating geometry-aware attention mechanisms, and establishing theoretical frameworks for stable training of hyperbolic foundation models. Graph-enhanced RAG systems and domain-specific applications in bioinformatics also show theoretical potential, representing important areas for future empirical validation.

\section{Conclusion}
Hyperbolic space can be regarded as a continuous tree, making it well-suited for modeling datasets with latent hierarchy layouts. HGL models extend the representation space to hyperbolic space and have achieved great success in graph data, especially with tree-like structures. 
In this survey, we present the technical details of HGL, including the methodologies, applications, challenges, and opportunities of HGL models. 
More specifically, we systematically categorize existing methods by learning paradigms and further examine their practical implementations across a wide range of application domains. 
We also identified several challenges that need to be overcome. To some extent, these challenges serve as guidelines for achieving the achievements of hyperbolic graph learning. Although many researchers are actively engaged in addressing these problems, we point to the numerous opportunities that still exist to
contribute to the development of this important, ever-evolving field.

\ifCLASSOPTIONcaptionsoff
  \newpage
\fi

\bibliographystyle{IEEEtran}
\bibliography{references}

\end{document}